\documentclass[journal]{IEEEtran}
\usepackage[T1]{fontenc}
\usepackage{amsmath, amsfonts}
\ifCLASSOPTIONcompsoc
    \usepackage[caption=false, font=normalsize, labelfont=sf, textfont=sf]{subfig}
\else
\usepackage[caption=false, font=footnotesize]{subfig}
\fi
\usepackage{cancel}
\usepackage[normalem]{ulem}

\usepackage{cite}
\usepackage{tabularx}
\usepackage{color}
\usepackage{scalerel}
\usepackage{booktabs}
\usepackage{bm}

\usepackage{multirow}
\newcolumntype{C}[1]{>{\centering\arraybackslash}p{#1}} % centering column type with fixed width
\newcolumntype{R}[1]{>{\raggedleft\arraybackslash}p{#1}} % right aligned column type with fixed width
\newcolumntype{L}[1]{>{\raggedright\arraybackslash}p{#1}} % left aligned column type with fixed width

\newcommand{\jmax}{j_\text{max}}
\newcommand{\hmax}{h_\text{max}}
\newcommand{\hbest}{h_\ell}
\newcommand{\jbest}{j_\ell}

\definecolor{shadecolor}{RGB}{150,255,150}

\newcommand{\padi}{pad_{i}}

\newcommand{\pad}[1]{pad_{#1}}

\newcommand{\MQUATLink}{\text{www.uni-kassel.de/go/mquat}}

\newcommand{\rout}{r_{\ell}}
\newcommand{\rin}{r_{\ell-1}}
\newcommand{\dout}{d_{\ell}}
\newcommand{\din}{d_{\ell-1}}

\newcommand{\SW}{x_{n + f \left\lfloor \frac{j}{k} \right\rfloor + (j \, \text{mod} \, k )}}

\usepackage{acro}

\DeclareAcronym{CNN}{
  short = CNN,
  long  = convolutional neural network
}

\DeclareAcronym{DNN}{
short=DNN,
long = deep neural network
}

\DeclareAcronym{NN}{
  short = NN,
  long  = neural network
}

\DeclareAcronym{PTQ}{
  short = PTQ,
  long  = post-training quantization
}

\DeclareAcronym{QAT}{
  short = QAT,
  long  = quantization-aware training
}

\DeclareAcronym{MAP}{
  short = mAP,
  long  = mean average precision
}

\DeclareAcronym{NLP}{
  short = NLP,
  long  = natural language processing
}

\DeclareAcronym{FPGA}{
  short = FPGA,
  long  = field-programmable gate array
}
\DeclareAcronym{GPU}{
  short = GPU,
  long  = graphics processing unit
}

\DeclareAcronym{TPU}{
  short = TPU,
  long  = tensor processing unit
}

\DeclareAcronym{CPU}{
  short = CPU,
  long  = Central processing unit
}

\DeclareAcronym{MAC}{
  short = MAC,
  long  = multiplication and accumulation
}

\DeclareAcronym{LUT}{
  short = LUT,
  long  = Lookup table
}

\DeclareAcronym{BNN}{
  short = BNN,
  long  = binary neural network
}

\DeclareAcronym{ASIC}{
  short = ASIC,
  long  = application-specific integrated circuit
}

\DeclareAcronym{IC}{
  short = IC,
  long  = integrated circuit
}

\DeclareAcronym{HWU}{
  short = HWU,
  long  = hardware unit
}

\begin{document}

%\title{Continuous-Flow Convolutional Neural Networks}
\title{Continuous-Flow Data-Rate-Aware\\ CNN Inference on FPGA}

\author{Tobias Habermann, Michael Mecik, Zhenyu Wang, César David Vera, Martin Kumm,~\IEEEmembership{Senior Member,~IEEE}, and Mario Garrido,~\IEEEmembership{Senior Member,~IEEE}  % <-this % stops a space
\thanks{Tobias Habermann, Michael Mecik, and Martin Kumm are with the Department of Applied Computer Science, Fulda University of Applied Sciences, 36037 Fulda, Germany, e-mails: tobias.habermann@informatik.hs-fulda.de, michael.mecik@informatik.hs-fulda.de, martin.kumm@informatik.hs-fulda.de}
\thanks{César David Vera and Mario Garrido are with the Department of Electronic Engineering, ETSI de Telecomunicaci\'on, Universidad Polit\'ecnica de Madrid, 28040 Madrid, Spain, e-mails: cesar.vera.moreno@alumnos.upm.es, mario.garrido@upm.es.}
\thanks{Zhenyu Wang is with the Department of Electrical Engineering, Linköping University, 58183 Linköping, Sweden, e-mail:zhewa394@student.liu.se}

\thanks{This work was supported in part by Comunidad de Madrid through the call "Ayudas de Estímulo a la Investigación de Jóvenes Doctores de la Universidad Politécnica de Madrid" under Project APOYO-JOVENES-21-TL23SB-116-I4FOMC; and in part by MCIN/AEI/10.13039/501100011033 and "ESF Investing in your future" under Grant RYC2018-025384-I; and in part by the Federal Ministry of Research, Technology and Space, Germany under Grant 05K25REA.}
}

\markboth{PREPRINT, COPYRIGHT IEEE, SEE HTTPS://DOI.ORG/10.1109/TCASAI.2026.3669843}%
{Shell \MakeLowercase{\textit{et al.}}: Continuous-Flow Convolutional Neural Networks}

\maketitle

\begin{abstract}

Among hardware accelerators for deep-learning inference, data flow implementations offer low latency and high throughput capabilities. In these architectures, each neuron is mapped to a dedicated hardware unit, making them well-suited for \ac{FPGA} implementation.
Previous unrolled implementations mostly focus on fully connected networks because of their simplicity, although it is well known that \acp{CNN} require fewer computations for the same accuracy.
When observing the data flow in \acp{CNN}, pooling layers and convolutional layers with a stride larger than one, the number of data at their output is reduced with respect to their input. This data reduction strongly affects the data rate in a fully parallel implementation, making hardware units heavily underutilized unless it is handled properly.
This work addresses this issue by analyzing the data flow of \acp{CNN} and presents a novel approach to designing data-rate-aware, continuous-flow CNN architectures. The proposed approach ensures a high hardware utilization close to 100\% by interleaving low data rate signals and sharing hardware units, as well as using the right parallelization to achieve the throughput of a fully parallel implementation.
The results show that a significant amount of the arithmetic logic can be saved, which allows implementing complex CNNs like MobileNet on a single FPGA with high throughput.

\end{abstract}

\begin{IEEEkeywords}
CNN, continuous-flow, parallel architecture, unrolled architecture
\end{IEEEkeywords}

\section{Introduction}
\label{sec:Intro}

\IEEEPARstart{C}{onvolutional} neural networks (CNNs) have proven to be powerful in various applications such as image classification~\cite{alexnet,resnet,mobilenet}, object detection~\cite{obj_det_survey,yolov1to8}, or \ac{NLP}~\cite{cnn_for_nlp}.
Various tasks, such as autonomous driving~\cite{selfdriving_challanges}, speech recognition~\cite{speechreq}, server/cloud computing, augmented or virtual reality~\cite{augreality},  and high energy physics \cite{engels_work}, require low-latency and high-throughput processing.
For these tasks, the use of \acp{NN} is challenging due to their high computational requirements, especially for applications that require real-time inference.
To meet these demands, inference accelerators are employed to enhance performance and efficiency.

Inference accelerators can be categorized into three types \cite[p.~734]{de_dinechin2024application}: general accelerators, data flow or stream architectures, and unrolled architectures.
General accelerators are usually implemented on custom \acp{IC} or application-specific \acp{IC} (ASICs) as those are fixed and can not be reconfigured like \acp{FPGA}. Hence, generic architectures like systolic arrays are used \cite{scye17} that use a matrix of processing engines to target large and medium-sized \acp{NN} \cite{tpu_eval,gpu_cnn_perf}.
%like \acp{GPU} and \acp{TPU},
Stream architectures \cite{umuroglu2017finn, duarte2018fast, venieris2016fpgaconvnet,main_ref} also target large and medium-sized \acp{NN} but provide a distinct hardware unit for each layer. They are network-specific and are more tailored and thus have typically lower resource and memory footprint than generic architectures \cite{de_dinechin2024application}.
Finally, unrolled architectures generally target small-sized \acp{NN} and are highly network-specialized and optimized \cite{umuroglu2020logicnets, andronic2023polylut, andronic2024neuralut, lou2024polylut, wang2019lutnet, ji2023fpqnet, dai2024kratos, main_ref, habermann2022hardware}.
In contrast to generic accelerators and stream architectures, unrolled architectures operate fully in parallel rather than sequentially, and most implementations of unrolled \cite{umuroglu2020logicnets, andronic2023polylut, andronic2024neuralut, lou2024polylut, habermann2022hardware} or partially unrolled architectures \cite{wang2019lutnet} are restricted to fully connected \acp{NN}.

To reduce the large resource requirements in fully unrolled \acp{NN}, many techniques have been proposed. They include quantization \cite{wang2019haq, miriyala2024mixed, dong2021hao}, pruning \cite{gao2023structural, gao2021network, hossain2023computational}, specific LUT-mappings \cite{umuroglu2020logicnets, andronic2023polylut, andronic2024neuralut, lou2024polylut, wang2019lutnet, ji2023fpqnet} for \acf{FPGA} implementations, and reducing the neuron inputs.
However, fully connected \acp{NN} still lack the scalability to implement larger models without exploding hardware costs.

An alternative to implement these larger models is to use \acp{CNN}. They have fewer computational requirements, as only a small data window is considered in the convolutional layer.
However, convolutional and pooling layers lead to changes in the data rate.
For example, in the common $2 \times 2$ max-pooling layer, the data rate at the output of the layer is one-fourth of the data rate at its input, since only one output is provided for each square of $2 \times 2$ values.

One solution to reach a full utilization of the hardware units in \acp{CNN} is to buffer data and share resources~\cite{tridgell2019unrolling, Zhenyu19MT}.
In~\cite{tridgell2019unrolling}, word- and bit-serial adders were used for layers with a reduced data rate.
This adaptation of the adders to the data rate reduces the hardware costs and ensures that the hardware is fully utilized.
However, it does not scale further to larger networks.

Another approach proposed in \cite{main_ref} considers a pixel-after-pixel architecture that is tailored to the data rate of the neural network.

In this approach, input data is fetched and then sorted into a buffer structure to allow the calculation of the current output of the layer. However, fetching and sorting the input data results in more overhead and requires more resources.

In this work, we present a new paradigm for the design of CNN accelerators that consists of an architecture that processes data in a continuous flow. 
A continuous flow architecture is able to process a continuous stream of data at a predefined input data rate
while the arithmetic units are matched such that they never have idle times or a lack of data.
This means that every arithmetic unit is computing every clock cycle.
Internal and output data rates are derived from the CNN model parameters.
The architecture allows to process data close to where the data is produced which keeps buffer sizes low.
This results in a higher utilization and thereby a smaller number of resources. 
In the context of the state-of-the-art, this approach bridges the gap between stream and unrolled architectures by permitting designs with different degrees of parallelization.
The main contributions of this paper are:
\begin{itemize}
\item A new paradigm for the design of continuous flow CNNs with different degrees of parallelization, covering the gap between stream and unrolled architectures.
\item A thorough analysis of CNN layers with respect to achieving continuous flow and minimizing idle times, including the explanation for interconnecting multiple layers for continuous flow, as well as an analysis of layer complexity depending on the parallelization.
\end{itemize}

The paper is organized as follows. In Section~\ref{sec:Background}, we review the main previous concepts about CNNs. In Section~\ref{sec:ContFlow}, we analyze the layers in CNNs from the point of view of continuous flow. In Section~\ref{sec:interleaving}, we explain how to build an entire continuous-flow CNN by combining different layers. In Section~\ref{sec:Comparison}, we analyze the complexity of the CNN in terms of hardware resources. In Section~\ref{sec:CommonCNNs}, we compare our designs to the fully parallel implementation of common CNN models. In Section~\ref{sec:Synthesis}, we present experimental results on FPGAs and compare them to the state-of-the-art. Finally, in Section~\ref{sec:Conclusions}, we summarize the main conclusions of the paper.

\section{Background}
\label{sec:Background}

\subsection{Convolutional Neural Networks}

\begin{figure}[t]
\centering
\includegraphics[width=\linewidth]{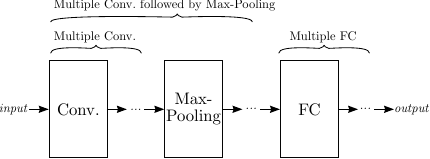}
\caption{The typical structure of a CNN.}
\label{fig.CNN_structure}
\end{figure}

\begin{figure*}[t]
\centering
\includegraphics[width=18 cm]{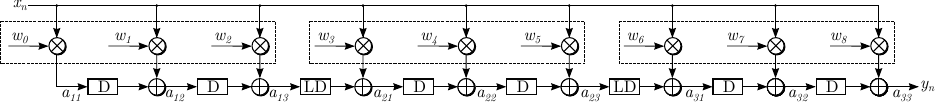}
\caption{KPU to calculate a $3 \times 3$ convolutional kernel.}
\label{fig.KPU1}
\end{figure*}

The typical structure of a CNN is shown in Fig.~\ref{fig.CNN_structure}. It consists of multiple blocks of convolutional layers with a max-pooling operation at the end of each block, with fully connected layers as the last layers of the model~\cite{cnns_review}.
The input and output of a CNN layer consist of multiple 2D feature maps forming a 3D tensor~\cite{cnns_review}.
This allows the CNN to process features in relation to their position in the tensor.
The three dimensions of the tensor are thereby split into the height, width, and channel dimension. Thereby, each channel has a unique feature map.
The core operation in CNNs is a sliding window operation applied on each input channel (feature map).

\subsection{Convolutional Layers}
In convolutional layers, each input value $x_n$ arrives at a discrete time step $n$.
The feature map has a size of $f \times f$ pixels.
Pixels are processed row by row, starting from the top-left corner.
According to this, the time of arrival of any input $x_n$ is related to the rows and columns of the feature map, as

\begin{equation}
\label{eq:calc_n}
n = r \cdot f + c \, ,
\end{equation}
where $r$ is the row index and $c$ is the column index in the feature map.
Using a grid of weights (kernel) with a shape of $k \times k$ the weighted sum of a sliding window from the input channel data $x$ is calculated as

\begin{equation}
\label{eq:calc_y_n}
y_n=\sum_{j=0}^{k^2-1} w_j \cdot x_{n + f \left\lfloor \frac{j}{k} \right\rfloor + (j \, \text{mod} \, k ) }    \, .
\end{equation}

The sum of all weighted sums of the same sliding window over all input channels, together with a bias, compose one output of an output channel (filter).
Each convolutional layer can have multiple filters.
Thus, each filter processes each input channel with different weights.

Intermediate accumulations of the summation are denoted as
\begin{equation}
\label{eq:calc_z_n}
z_{n,i}=\sum_{j=0}^{i} w_j \cdot x_{n + f \left\lfloor \frac{j}{k} \right\rfloor + (j \, \text{mod} \, k )}   \, ,
\end{equation}

where $z_{n,i}$ is the result of accumulating the first products until and including index $i$ that lead to the output $y_n$. Therefore, the output corresponds to
\begin{equation}
\label{eq:calc_y_n_2}
y_n = z_{n,k^2-1} \, .
\end{equation}

When $y_n$ describes the result of the operation applied on the sliding window with $x_n$ in the upper-left corner, the last $k-1$ rows and columns of the feature map would result in a sliding window that is outside the feature map.
Thus, valid outputs are defined as
\begin{equation}
\label{eq:valid_y_n}
y_n \, \text{is} \, \text{valid} \, \forall \, n = r \cdot f + c \, | \, r,c \in \{ 0, ..., f - k  \}   \, .
\end{equation}

\subsection{Pooling Layers}
Pooling layers also apply a sliding window operation.
Each input channel is processed using a pooling operation and defines the output of an output channel.
So, pooling layers always have as many input channels as they have output channels.
For example, for a max-pooling layer, the sliding-window operation would be

\begin{equation}
\label{eq:max_pooling_eq}
y_{n} = \max_{j=0}^{k^2-1}\left \{ \SW \right \} \, ,
\end{equation}
where $\text{max}$ can be replaced by other pooling operations.
The most common pooling operations are max and average pooling, where the maximum value or the average is calculated for each sliding window.

\subsection{Fully Connected Layers}
Fully connected layers do not apply a sliding window operation.
Each neuron in a fully connected layer has an individual weight for each input feature.
The output $y_n$ of the $n$-th neuron for the input channel data $x$ is the weighted sum of all inputs and can be described as

\begin{equation}
\label{eq:fc_calc}
y_{n} = \sum_{j=0}^{f^2-1}x_j \cdot w_j  \, .
\end{equation}

For simplification, the fully connected layer operation can be described with a sliding window operation.
This criterion can be satisfied by setting $k=f$.
Each neuron can then be represented as a filter and
the fully connected layer can be treated like a convolutional layer.

\begin{table}[t]
\centering
\caption{Timing of the KPU in Fig.~\ref{fig.KPU1} for a $5 \times 5$ feature map with a $3 \times 3$ kernel.}
\label{td.kpu1}
\begin{tabular}{cccccccc}
\toprule
\textbf{t} & $\bm{x_n}$ & $\bm{a_{11}}$  & $\bm{a_{13}}$ & $\bm{a_{21}}$  & $\bm{a_{23}}$ & $\bm{a_{31}}$ & $\bm{y_n}$                         \\
\midrule
0 & $x_{0}$ & $z_{0,0}$ & - & - & - & - & - \\
1 & $x_{1}$ & $z_{1,0}$ & - & - & - & - & - \\
2 & $x_{2}$ & $z_{2,0}$ & $z_{0,2}$ & - & - & - & - \\
3 & $x_{3}$ & - & $z_{1,2}$ & - & - & - & - \\
4 & $x_{4}$ & - & $z_{2,2}$ & - & - & - & - \\
5 & $x_{5}$ & $z_{5,0}$ & - & $z_{0,3}$ & - & - & - \\
6 & $x_{6}$ & $z_{6,0}$ & - & $z_{1,3}$ & - & - & - \\
7 & $x_{7}$ & $z_{7,0}$ & $z_{5,2}$ & $z_{2,3}$ & $z_{0,5}$ & - & - \\
8 & $x_{8}$ & - & $z_{6,2}$ & - & $z_{1,5}$ & - & - \\
9 & $x_{9}$ & - & $z_{7,2}$ & - & $z_{2,5}$ & - & - \\
10 & $x_{10}$ & $z_{10,0}$ & - & $z_{5,3}$ & - & $z_{0,6}$ & - \\
11 & $x_{11}$ & $z_{11,0}$ & - & $z_{6,3}$ & - & $z_{1,6}$ & - \\
12 & $x_{12}$ & $z_{12,0}$ & $z_{10,2}$ & $z_{7,3}$ & $z_{5,5}$ & $z_{2,6}$ & $y_{0}$ \\
13 & $x_{13}$ & - & $z_{11,2}$ & - & $z_{6,5}$ & - & $y_{1}$ \\
14 & $x_{14}$ & - & $z_{12,2}$ & - & $z_{7,5}$ & - & $y_{2}$ \\
15 & $x_{15}$ & - & - & $z_{10,3}$ & - & $z_{5,6}$ & - \\
16 & $x_{16}$ & - & - & $z_{11,3}$ & - & $z_{6,6}$ & - \\
17 & $x_{17}$ & - & - & $z_{12,3}$ & $z_{10,5}$ & $z_{7,6}$ & $y_{5}$ \\
18 & $x_{18}$ & - & - & - & $z_{11,5}$ & - & $y_{6}$ \\
19 & $x_{19}$ & - & - & - & $z_{12,5}$ & - & $y_{7}$ \\
20 & $x_{20}$ & - & - & - & - & $z_{10,6}$ & - \\
21 & $x_{21}$ & - & - & - & - & $z_{11,6}$ & - \\
22 & $x_{22}$ & - & - & - & - & $z_{12,6}$ & $y_{10}$ \\
23 & $x_{23}$ & - & - & - & - & - & $y_{11}$ \\
24 & $x_{24}$ & - & - & - & - & - & $y_{12}$ \\
\bottomrule
\end{tabular}
\end{table}

\begin{figure}[t]
  \centering
  \subfloat[Without padding.]{
    \includegraphics[width=0.3\linewidth]{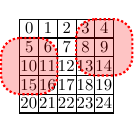}
    \label{fig.dataorder_nopad}
  }
  \hspace{0.1\linewidth}   % maximize separation between the subfigures
  \subfloat[With padding.]{
    \includegraphics[width=0.3\linewidth]{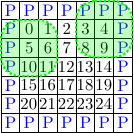}
    \label{fig.dataorder_pad}
  }
\caption{Diagrams to analyze continuous flow with and without padding.}
\label{fig.dataorder_combo}
\end{figure}

\section{Continuous Flow in CNN Layers}
\label{sec:ContFlow}

In this section, we analyze convolutional, max-pooling, and fully connected layers from the point of view of achieving continuous flow.
%The continuous flow at the input of the layers may or may not lead to continuous flow at the output under certain circumstances.
Once these layers are modelled, in Section~\ref{sec:interleaving} we show how to combine the different data rates that the layers of the CNN produce to achieve continuous flow, which allows for building an entire continuous-flow CNN.

In order to model continuous flow, we consider that each input pixel at layer $l$ consists of $\din$ features and produces one output pixel consisting of $\dout$ features.
%This would lead to an output data rate of $\frac{\dout}{\din}$.
The input data rate, $\rin$, is defined as the number of input features per clock cycle. Thereby, it is a factor that controls how fast the output data can be generated.
Finally, the stride of the operation also affects the number of output features per clock cycle by only keeping $1/s^2$ of the output pixels.
The output data rate, $\rout$, can thereby be calculated as
\begin{equation}
\label{eq.rout}
\rout = \frac{\dout \cdot \rin}{\din \cdot s^2}.
\end{equation}
This equation holds for any type of layer.

\subsection{Continuous Flow in Convolutional Layers}
\label{sec:ConvLayers}
The component that implements the convolution in a CNN is called kernel processing unit (KPU).
An example of a simple KPU that calculates a $3 \times 3$ convolution is shown in Fig.~\ref{fig.KPU1}.
It uses the transposed form \cite{kfmzm13}, which allows for an efficient implementation of the multiplier and post-adder of a digital signal processor (DSP) block in common FPGAs.
%Furthermore, the transposed form allows a continuous flow at the input.
This KPU circuit implements a classic row buffering scheme \cite{b11} and calculates one $3 \times 3$ convolution per clock cycle by using 9 general multipliers ($\otimes$), 8 adders ($\oplus$), 6 registers ($\text{D}$) and 2 line buffers ($\text{LD}$).

Fig.~\ref{fig.dataorder_nopad} shows a $5 \times 5$ feature map. The timing in Table~\ref{td.kpu1} shows how the KPU in Fig.~\ref{fig.KPU1} processes this feature map considering a stride of $s = 1$.
Invalid values are indicated with (-).
Each set of multipliers, marked with a dashed box in Fig.~\ref{fig.KPU1}, computes one row of the sliding window.
The line buffer in the preceding set delays the data by $L = f - k + 1$.
The timing shows that the KPU achieves a continuous flow at the input, but does not achieve a continuous flow at the output because there are invalid outputs in between valid ones. For instance, the output at $t=15$ and $t=16$ in Table~\ref{td.kpu1} are invalid and correspond to the sliding windows highlighted in Fig.~\ref{fig.dataorder_nopad}, which are outside the feature map.

\subsection{Continuous Flow with Padding in Convolutional Layers}

\begin{figure*}[t]
    \centering
    \includegraphics{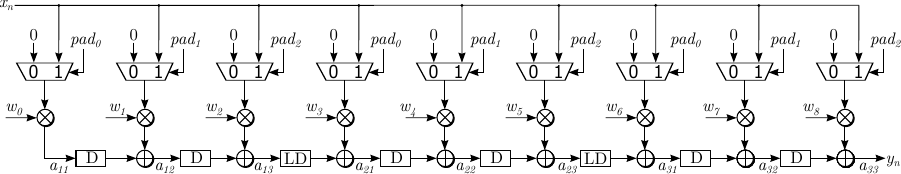}
	\caption{A KPU with added zero-padding functionality.}
    \label{fig.kpu1_pad}
\end{figure*}

Padding is a well-known technique in CNNs to preserve spatial dimensions and edge information. When padding is used in the convolutional layer, it is also possible to achieve a continuous flow at the input and the output of the KPU. In this work, we focus on zero-padding because it is the most common and the most simple padding type.
It can be observed that the conventional strategy of feeding zeros into the KPU at the start and end of each row breaks the continuous flow at the input by interrupting the valid inputs with explicit zero padding.

By contrast, we suggest to use implicit zero padding by setting specific multipliers of the KPU to zero, depending on the column index of the input pixel. Thus, the data order does not change and the input and output are continuous.
To illustrate this idea, Fig.~\ref{fig.dataorder_pad} shows the data order for a $5 \times 5$ feature map with zero-padding $p=1$, where $p$ represents the number of padded zeros at the sides of the feature map.
Fig.~\ref{fig.dataorder_pad} also highlights the same two sliding windows that were highlighted in the Fig.~\ref{fig.dataorder_nopad}.
However, in Fig.~\ref{fig.dataorder_pad}, it can be seen that both sliding windows are now contained inside the feature map, leading to continuous flow at the output.
In general, adding padding to convolutional layers leads to continuous flow at the output as long as $p=\frac{k-1}{2}$. The definition of valid outputs for convolutional layers in (\ref{eq:valid_y_n}) is then adapted to 
%The definition of valid outputs is adapted by adding a padding term that extends the set of valid outputs to
\begin{equation}\label{eq:valid_y_n_pad}
y_n \, \text{is} \, \text{valid} \, \forall \, n = r \cdot f + c \, | \, r,c \in \{ 0, ..., f - k + 2p \}  \, .
\end{equation}

To implement implicit zero-padding in the KPU, multiplexers are added in front of the general multipliers, as shown in Fig.~\ref{fig.kpu1_pad}.
Each of the three multiplier columns has its own select signal that depends on the current column index of the input pixel.
Each multiplexer can be simplified as a bitwise AND gate.
This allows for disabling the calculation of the selected columns for each input pixel by setting the weights for the column to zero.
All padding select signals, $\padi$, are derived from the column index $c$ of the current input pixel.
There are $k$ select signals in total for each KPU. The equation for all $\padi$ signals as a function of the column is % shown in~\ref{eq:padding_select}.
\label{eq:padding_select}
\begin{equation}
\padi(c) =
\begin{cases}
0 & \text{if } c \geq f - p + i, \\
0 & \text{if } c < p - k + i + 1, \\
1 & \text{otherwise}.
\end{cases}
\end{equation}
In general, if the current input pixel is located at the far left or the far right of the feature map, select signals corresponding to the opposite side are set to zero.
For example, with $k=3, p=1, f=5$ there are three select signals, $\pad{0}$, $\pad{1}$, and $\pad{2}$. When $c=0$, the current input pixel is at the far left, which makes $\pad{2}$ become zero because $c<p-k+i+1$ is true, thereby padding the far-right column with zero.
The opposite applies when $c=4$, where $\pad{0}$ becomes zero because $c \geq f - p + i$ is true. With the $\padi$ signals, the KPU can now mask the overlapping left and right columns of invalid sliding windows, as shown in Fig.~\ref{fig.dataorder_pad}.

\newcommand{\colwid}{6.09mm}
\begin{table}[t]
\centering
\caption{Timing of the KPU in Fig.~\ref{fig.kpu1_pad} applied on a $5 \times 5$ feature map with padding $p=1$. The padding column describes all three padding select signals as a tuple $(\pad{0}, \pad{1}, \pad{2})$.}
\label{td.kpu1_pad}
%\begin{scriptsize}
\centering
\begin{tabular}{C{2.8mm}C{4.5mm}C{7.5mm}C{\colwid}C{\colwid}C{\colwid}C{\colwid}C{\colwid}C{5mm}}
\toprule
\textbf{t} & $\bm{x_n}$ & \textbf{Pad} & $\bm{a_{11}}$ & $\bm{a_{13}}$ & $\bm{a_{21}}$ & $\bm{a_{23}}$ & $\bm{a_{31}}$ & $\bm{y_n}$               \\ \midrule
0             &   0             &   -             &   $z_{0, 0}$    &   -             &   -             &   -             &   -             &   -           \\
1             &   0             &   -             &   $z_{1, 0}$    &   -             &   -             &   -             &   -             &   -           \\
2             &   0             &   -             &   $z_{2, 0}$    &   $z_{0, 2}$    &   -             &   -             &   -             &   -           \\
3             &   0             &   -             &   $z_{3, 0}$    &   $z_{1, 2}$    &   -             &   -             &   -             &   -           \\
4             &   0             &   -             &   $z_{4, 0}$    &   $z_{2, 2}$    &   -             &   -             &   -             &   -           \\
5             &   0             &   -             &   $z_{5, 0}$    &   $z_{3, 2}$    &   $z_{0, 3}$    &   -             &   -             &   -           \\
6             &   $x_{0}$       &   (1,1,0)       &   $z_{6, 0}$    &   $z_{4, 2}$    &   $z_{1, 3}$    &   -             &   -             &   -           \\
7             &   $x_{1}$       &   (1,1,1)       &   $z_{7, 0}$    &   $z_{5, 2}$    &   $z_{2, 3}$    &   $z_{0, 5}$    &   -             &   -           \\
8             &   $x_{2}$       &   (1,1,1)       &   $z_{8, 0}$    &   $z_{6, 2}$    &   $z_{3, 3}$    &   $z_{1, 5}$    &   -             &   -           \\
9             &   $x_{3}$       &   (1,1,1)       &   $z_{9, 0}$    &   $z_{7, 2}$    &   $z_{4, 3}$    &   $z_{2, 5}$    &   -             &   -           \\
10            &   $x_{4}$       &   (0,1,1)       &   $z_{10, 0}$   &   $z_{8, 2}$    &   $z_{5, 3}$    &   $z_{3, 5}$    &   $z_{0, 6}$    &   -           \\
11            &   $x_{5}$       &   (1,1,0)       &   $z_{11, 0}$   &   $z_{9, 2}$    &   $z_{6, 3}$    &   $z_{4, 5}$    &   $z_{1, 6}$    &   -           \\
12            &   $x_{6}$       &   (1,1,1)       &   $z_{12, 0}$   &   $z_{10, 2}$   &   $z_{7, 3}$    &   $z_{5, 5}$    &   $z_{2, 6}$    &   $y_{0}$     \\
13            &   $x_{7}$       &   (1,1,1)       &   $z_{13, 0}$   &   $z_{11, 2}$   &   $z_{8, 3}$    &   $z_{6, 5}$    &   $z_{3, 6}$    &   $y_{1}$     \\
14            &   $x_{8}$       &   (1,1,1)       &   $z_{14, 0}$   &   $z_{12, 2}$   &   $z_{9, 3}$    &   $z_{7, 5}$    &   $z_{4, 6}$    &   $y_{2}$     \\
15            &   $x_{9}$       &   (0,1,1)       &   $z_{15, 0}$   &   $z_{13, 2}$   &   $z_{10, 3}$   &   $z_{8, 5}$    &   $z_{5, 6}$    &   $y_{3}$     \\
16            &   $x_{10}$      &   (1,1,0)       &   $z_{16, 0}$   &   $z_{14, 2}$   &   $z_{11, 3}$   &   $z_{9, 5}$    &   $z_{6, 6}$    &   $y_{4}$     \\
17            &   $x_{11}$      &   (1,1,1)       &   $z_{17, 0}$   &   $z_{15, 2}$   &   $z_{12, 3}$   &   $z_{10, 5}$   &   $z_{7, 6}$    &   $y_{5}$     \\
18            &   $x_{12}$      &   (1,1,1)       &   $z_{18, 0}$   &   $z_{16, 2}$   &   $z_{13, 3}$   &   $z_{11, 5}$   &   $z_{8, 6}$    &   $y_{6}$     \\
19            &   $x_{13}$      &   (1,1,1)       &   $z_{19, 0}$   &   $z_{17, 2}$   &   $z_{14, 3}$   &   $z_{12, 5}$   &   $z_{9, 6}$    &   $y_{7}$     \\
20            &   $x_{14}$      &   (0,1,1)       &   $z_{20, 0}$   &   $z_{18, 2}$   &   $z_{15, 3}$   &   $z_{13, 5}$   &   $z_{10, 6}$   &   $y_{8}$     \\
21            &   $x_{15}$      &   (1,1,0)       &   $z_{21, 0}$   &   $z_{19, 2}$   &   $z_{16, 3}$   &   $z_{14, 5}$   &   $z_{11, 6}$   &   $y_{9}$     \\
22            &   $x_{16}$      &   (1,1,1)       &   $z_{22, 0}$   &   $z_{20, 2}$   &   $z_{17, 3}$   &   $z_{15, 5}$   &   $z_{12, 6}$   &   $y_{10}$    \\
23            &   $x_{17}$      &   (1,1,1)       &   $z_{23, 0}$   &   $z_{21, 2}$   &   $z_{18, 3}$   &   $z_{16, 5}$   &   $z_{13, 6}$   &   $y_{11}$    \\
24            &   $x_{18}$      &   (1,1,1)       &   $z_{24, 0}$   &   $z_{22, 2}$   &   $z_{19, 3}$   &   $z_{17, 5}$   &   $z_{14, 6}$   &   $y_{12}$    \\
25            &   $x_{19}$      &   (0,1,1)       &   -             &   $z_{23, 2}$   &   $z_{20, 3}$   &   $z_{18, 5}$   &   $z_{15, 6}$   &   $y_{13}$    \\
26            &   $x_{20}$      &   (1,1,0)       &   -             &   $z_{24, 2}$   &   $z_{21, 3}$   &   $z_{19, 5}$   &   $z_{16, 6}$   &   $y_{14}$    \\
27            &   $x_{21}$      &   (1,1,1)       &   -             &   -             &   $z_{22, 3}$   &   $z_{20, 5}$   &   $z_{17, 6}$   &   $y_{15}$    \\
28            &   $x_{22}$      &   (1,1,1)       &   -             &   -             &   $z_{23, 3}$   &   $z_{21, 5}$   &   $z_{18, 6}$   &   $y_{16}$    \\
29            &   $x_{23}$      &   (1,1,1)       &   -             &   -             &   $z_{24, 3}$   &   $z_{22, 5}$   &   $z_{19, 6}$   &   $y_{17}$    \\
30            &   $x_{24}$      &   (0,1,1)       &   -             &   -             &   -             &   $z_{23, 5}$   &   $z_{20, 6}$   &   $y_{18}$    \\
31            &   0             &   -             &   $z_{0, 0}$    &   -             &   -             &   $z_{24, 5}$   &   $z_{21, 6}$   &   $y_{19}$    \\
32            &   0             &   -             &   $z_{1, 0}$    &   -             &   -             &   -             &   $z_{22, 6}$   &   $y_{20}$    \\
33            &   0             &   -             &   $z_{2, 0}$    &   $z_{0, 2}$    &   -             &   -             &   $z_{23, 6}$   &   $y_{21}$    \\
34            &   0             &   -             &   $z_{3, 0}$    &   $z_{1, 2}$    &   -             &   -             &   $z_{24, 6}$   &   $y_{22}$    \\
35            &   0             &   -             &   $z_{4, 0}$    &   $z_{2, 2}$    &   -             &   -             &   -             &   $y_{23}$    \\
36            &   0             &   -             &   $z_{5, 0}$    &   $z_{3, 2}$    &   $z_{0, 3}$    &   -             &   -             &   $y_{24}$    \\
\bottomrule
\end{tabular}
%\end{scriptsize}
\end{table}

Table~\ref{td.kpu1_pad} shows the timing for the KPU with a stride $s=1$ and a padding $p=1$ in Fig.~\ref{fig.kpu1_pad} used to process a $5 \times 5$ feature map.
It can be seen that this KPU achieves continuous flow both at the input and at the output.
For the very first row, the KPU has to be fed with zeros %for $\frac{(f+1)(k-1)}{2}$ clock cycles because
to implement
the padding of the top row.
Likewise, after all inputs are received, %another $\frac{(f+1)(k-1)}{2}$ clock cycles are used to
zeros are fed into the KPU to implement the padding at the bottom row and generate the last valid outputs.
This padding of the bottom row is then reused as a padding of the top for the next feature map.

To summarize, adding multiplexers to the KPU allows achieving a continuous flow for $s=1$ by introducing zero padding where $p=\frac{k-1}{2}$.

\subsection{Continuous Flow With Stride In Convolutional Layers}

With a stride $s = 1$, the output of the KPU is considered valid when the corresponding sliding window is fully contained in the feature map.
This is already included in (\ref{eq:valid_y_n_pad}).
With a stride $s > 1$, outputs are skipped where the row and column index are not multiples of $s$.
Therefore, the validity of $y_n$ has to be adapted to
\begin{equation}
\label{eq:valid_y_n_stride}
y_n \, \text{is} \, \text{valid} \, \forall \, n = r \cdot f + c \, | \, r,c \in \{ 0, s, 2s, ..., f - k + 2p \}.
\end{equation}
When processing pixel after pixel, a stride $s > 1$ leads inherently to a non-continuous output. We will explain how to handle this case in Section~\ref{sec:interleaving_conv}.

\begin{figure}[t]
    \centering
    \includegraphics{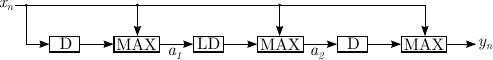}
	\caption{An example for a $2 \times 2$ max-pooling operation.}
    \label{fig.2x2max_pool}
\end{figure}

\subsection{Continuous Flow in Pooling Layers}
A typical max-pooling operation is implemented using the pooling processing unit (PPU) for the example of a $2 \times 2$ kernel and a stride of $s = 2$ is shown in Fig.~\ref{fig.2x2max_pool}.
It uses 2 registers (D), 3 max units that calculate the maximum among their two inputs (MAX), and one line buffer (LD). The circuit works similarly to the KPU, where the line buffer ensures that the correct four pixels are compared to each other accordingly to the sliding window.

With the nature of pooling layers to reduce the feature map size, the stride of pooling layers is generally larger than one, which leads to a non-continuous output with a data rate of only $1/s^2$. This issue has to be addressed in the CNN implementation with the goal of achieving continuous flow and is discussed later in Section~\ref{sec:interleaving_pooling}.

\subsection{Continuous Flow in Fully Connected Layers}

Fully connected layers are typically the last layers in a CNN.
The three-dimensional tensor containing all feature maps is flattened into a single dimension.
%The pixels lose their spacial context and are called features.
Each neuron in the fully connected layer is connected to all input features.
With a continuous flow of input features, the first valid output of a fully connected layer can only be obtained after all inputs are received.
%The last layer has the lowest data rate and has to be designed with this in mind.

\begin{figure}[t]
    \centering
    \includegraphics[width=1.0\linewidth]{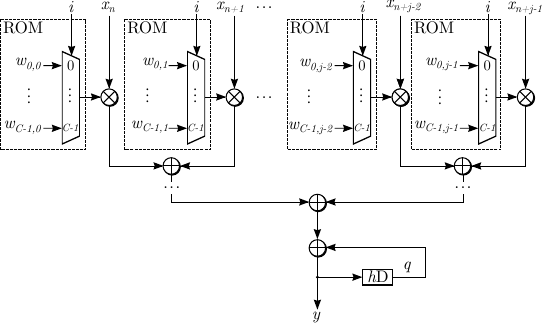}
	\caption{An example of an FCU that calculates $h$ different neurons with $j$ inputs.}
    \label{fig.fc}
\end{figure}

A fully connected layer is implemented using the fully connected unit (FCU) shown in Fig.~\ref{fig.fc}. This FCU implements $h$ neurons of the fully connected layer.
The circuit holds $j$ input pixels for $h$ clock cycles and switches between the weights every clock cycle.
The multiplexers that are used to switch between weights are implemented as read-only memory (ROM). Because the FCU processes multiple inputs in parallel, the weight tensor $w$ must also include an index corresponding to each input.
In each of the $h$ clock cycles, the partial sum of the $j$ weighted inputs is calculated for the corresponding neuron and added to the previous partial sum of the corresponding neuron in the buffer.
After $h$ cycles, the next $j$ inputs are loaded. When the last $j$ inputs are processed, the FCU produces $h$ valid outputs for the corresponding neurons.

\renewcommand{\colwid}{0.75cm}

\begin{table}[t]
\centering
\caption{Timing of the FCU in Fig.~\ref{fig.fc} with $h=5$ and $j=4$.}
\label{td.FC}
\begin{tabular}{C{0.65cm}C{0.2cm}C{0.595cm}C{0.595cm}C{0.595cm}C{0.595cm}C{0.48cm}C{0.48cm}}
\toprule
\textbf{t}   &    $\bm{n}$   & $\bm{w_{i,0}}$ & $\bm{w_{i,1}}$ & $\bm{w_{i,2}}$ & $\bm{w_{i,3}}$ & $\bm{q}$ & $\bm{y}$ \\
\midrule
0 & $0$ & $w_{0,0}$ & $w_{0,1}$ & $w_{0,2}$ & $w_{0,3}$ & 0 & $z_{0,3}$ \\
1 & $0$ & $w_{1,0}$ & $w_{1,1}$ & $w_{1,2}$ & $w_{1,3}$ & 0 & $z_{1,3}$ \\
2 & $0$ & $w_{2,0}$ & $w_{2,1}$ & $w_{2,2}$ & $w_{2,3}$ & 0 & $z_{2,3}$ \\
3 & $0$ & $w_{3,0}$ & $w_{3,1}$ & $w_{3,2}$ & $w_{3,3}$ & 0 & $z_{3,3}$ \\
4 & $0$ & $w_{4,0}$ & $w_{4,1}$ & $w_{4,2}$ & $w_{4,3}$ & 0 & $z_{4,3}$ \\
5 & $4$ & $w_{5,0}$ & $w_{5,1}$ & $w_{5,2}$ & $w_{5,3}$ & $z_{0,3}$ & $y_{0}$ \\
6 & $4$ & $w_{6,0}$ & $w_{6,1}$ & $w_{6,2}$ & $w_{6,3}$ & $z_{1,3}$ & $y_{1}$ \\
7 & $4$ & $w_{7,0}$ & $w_{7,1}$ & $w_{7,2}$ & $w_{7,3}$ & $z_{2,3}$ & $y_{2}$ \\
8 & $4$ & $w_{8,0}$ & $w_{8,1}$ & $w_{8,2}$ & $w_{8,3}$ & $z_{3,3}$ & $y_{3}$ \\
9 & $4$ & $w_{9,0}$ & $w_{9,1}$ & $w_{9,2}$ & $w_{9,3}$ & $z_{4,3}$ & $y_{4}$ \\
\bottomrule
\end{tabular}
\end{table}
With each FCU implementing $h$ neurons and by processing $j$ inputs in one clock cycle, the FCUs have to switch between different weight configurations for the multipliers. The number of configurations is calculated by:
\begin{equation}
\label{eq:calc_configs_fc}
C=h \cdot \frac{d_{\ell-1}}{j} \, .
\end{equation}
The timing of the FCU in Fig.~\ref{fig.fc} when calculating $h=5$ neurons with $j=4$ inputs for 8 input features in total is shown in Table~\ref{td.FC}. With 8 input features in total, the FCU calculates all outputs after switching the inputs of the FCU twice.
%The first 5 clock cycles process the first 4 inputs, and the last 5 clock cycles process the last 4 inputs.
At each of the first 5 clock cycles, a partial sum of products for one of the 5 neurons is calculated. The signal $q$ for the buffer $hD$ holds the accumulated result of the current neuron.
The index $i$ is incremented each clock cycle and is reset after all 8 inputs are processed for all 5 neurons.

To implement a fully connected layer, the goal is then to find the number of inputs $\jbest$ and neurons $\hbest$ for the FCUs that satisfy the input data rate criteria.
To do so, the input data rate $\rin$ is interpreted as
\begin{equation}
\label{eq.biggest_fcu}
\rin = \frac{\jmax}{\hmax} ,
\end{equation}
where $\jmax$ inputs are processed over $\hmax$ clock cycles.
With each FCU processing $\jbest=\jmax$ inputs, the number of neurons for each FCU, $\hbest$, can be derived from the greatest divisor of $\dout$ that is smaller or equal to $\hmax$, i.e., 
\begin{equation}
\label{eq.best_fcu}
\hbest = \text{max}(\{ h \in \mathbb{N} \, | \, h \text{ divides } \dout, h \leq \hmax \})  \, .
\end{equation}
This allows for high utilization of all FCUs.
When implementing FCUs, the adder for accumulation could have more pipeline stages than $h$.
However, the buffer depth of the FCU has to be $h$, as shown in Fig.~\ref{fig.fc}, to implement $h$ different summations.

If the number of pipeline stages is larger than $h$, the design fails.
We solve this issue by ensuring that $h$ is big enough by introducing data aggregation.
By aggregating inputs $a$ times, the definition for $\jmax$ and $\hmax$ change to

\begin{equation}
\label{eq.biggest_fcu_aggr}
\rin = \frac{a \cdot \jmax}{a \cdot \hmax} \, .
\end{equation}

\begin{figure}[t]
    \centering
    \includegraphics{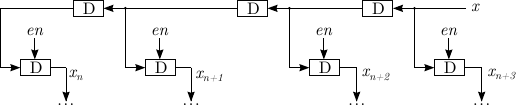}
	\caption{The data aggregation circuit that is connected upstream of the FCU to aggregate 4 inputs.}
    \label{fig.fcu_aggregation_example}
\end{figure}

This also reduces the number of FCUs needed because each FCU calculates multiple neurons.
For example, with $\rin=1$, and $\din=8$, each FCU could only process one input and calculate one neuron. With $a=4$ each FCU can process 4 inputs and calculate 4 neurons.
The aggregation circuit that is connected upstream of the FCU and aggregates 4 inputs is shown in Fig.~\ref{fig.fcu_aggregation_example}.
The timing after aggregation is shown in Table~\ref{td.FCU_h4_j4}. In this case, the FCU produces a valid output after 9 clock cycles.
Without aggregation, the FCU would produce a valid output after 8 clock cycles.
So the additional delay for aggregation is small.

\begin{table}[t]
\centering
\caption{Timing of an FCU with $h=4$ and $j=4$ with $\din=8$.}
\label{td.FCU_h4_j4}
\begin{tabular}{C{0.7cm}C{1.0cm}C{0.6cm}C{0.6cm}C{0.6cm}C{0.6cm}C{0.5cm}C{0.5cm}}
\toprule
\textbf{t}         & $\bm{x}$ & $\bm{w_{i,0}}$ & $\bm{w_{i,1}}$ & $\bm{w_{i,2}}$ & $\bm{w_{i,3}}$ & $\bm{q}$ & $\bm{y_{n}}$ \\
\midrule
0 & (-,-,-,-)  & - & - & - & - & 0 & - \\
1 & (-,-,-,0)  & - & - & - & - & 0 & - \\
2 & (-,-,0,1)  & - & - & - & - & 0 & - \\
3 & (-,0,1,2)  & - & - & - & - & 0 & - \\
4 & (0,1,2,3)  & $w_{0,0}$ & $w_{0,1}$ & $w_{0,2}$ & $w_{0,3}$ & 0 & $z_{0,3}$ \\
5 & (0,1,2,3)  & $w_{1,0}$ & $w_{1,1}$ & $w_{1,2}$ & $w_{1,3}$ & 0 & $z_{1,3}$ \\
6 & (0,1,2,3)  & $w_{2,0}$ & $w_{2,1}$ & $w_{2,2}$ & $w_{2,3}$ & 0 & $z_{2,3}$ \\
7 & (0,1,2,3)  & $w_{3,0}$ & $w_{3,1}$ & $w_{3,2}$ & $w_{3,3}$ & 0 & $z_{3,3}$ \\
8 & (4,5,6,7)  & $w_{4,0}$ & $w_{4,1}$ & $w_{4,2}$ & $w_{4,3}$ & $z_{0,3}$ & $y_{0}$ \\
9 & (4,5,6,7)  & $w_{5,0}$ & $w_{5,1}$ & $w_{5,2}$ & $w_{5,3}$ & $z_{1,3}$ & $y_{1}$ \\
10 & (4,5,6,7) & $w_{6,0}$ & $w_{6,1}$ & $w_{6,2}$ & $w_{6,3}$ & $z_{2,3}$ & $y_{2}$ \\
11 & (4,5,6,7) & $w_{7,0}$ & $w_{7,1}$ & $w_{7,2}$ & $w_{7,3}$ & $z_{3,3}$ & $y_{3}$ \\
\bottomrule
\end{tabular}
\end{table}

\section{Building an Entire Continuous-Flow CNN}
\label{sec:interleaving}

This section shows how to build a continuous-flow CNN. This involves two key design aspects. First, the data rates and hardware resources of all the layers are adjusted so that they can keep the data flow through the entire CNN. Second, idle times caused by invalid outputs of the layers are reduced so that processing units are not underutilized.

\subsection{Running Example}

To illustrate the concepts that we present in this section, a running example is used. 
The parameters of the running example are shown in Table~\ref{td.datarates_example_breakdown}.

It consists of a CNN with five layers, where C1 and C2 are convolutional layers, P1 and P2 are max-pooling layers, and F1 is a fully connected layer.
%The example model is implemented according to the presented continuous-flow architecture and consists of several layers.
\begin{figure*}[t]
    \centering
    \includegraphics[width=18cm]{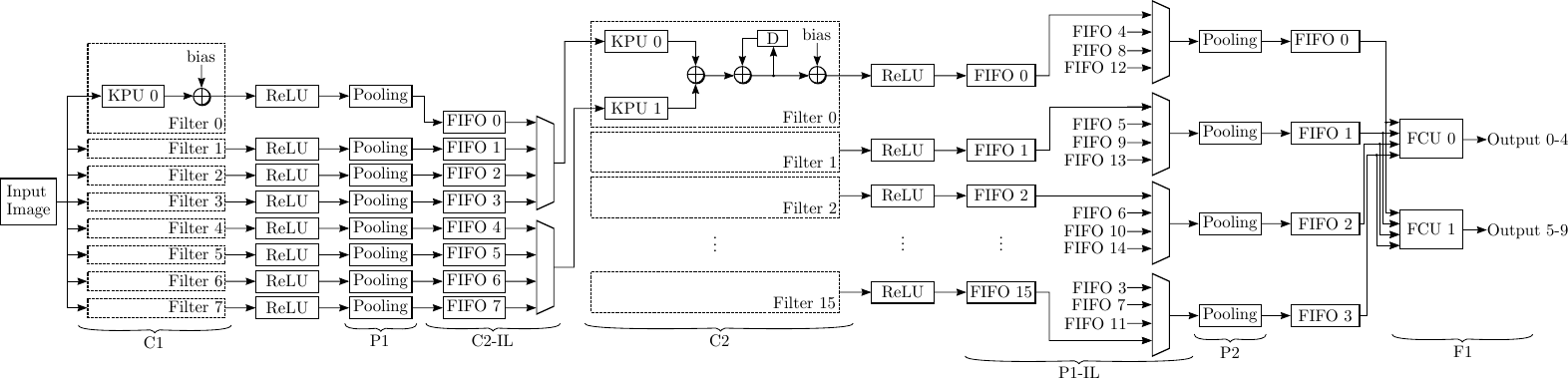}
	\caption{Continuous-flow CNN architecture that results from the implementation of the running example.}
    \label{fig.running_example}
\end{figure*}
\begin{table*}[t]
\centering
\caption{The structure and analysis of the running example for each layer.}
\label{td.datarates_example_breakdown}
\begin{tabular}{L{6.99mm}C{11.9mm} C{2.64mm}  C{1.67mm}C{1.45mm}C{1.64mm}C{3.0mm} C{4.05mm}C{5.14mm}     C{15.75mm}   C{5.7mm}C{5.6mm}C{6mm}C{7mm}C{7mm}C{6mm}C{6mm}C{6mm}}

\toprule
\textbf{Layer} & \textbf{Input} &     $\bm{f}$ &     $\bm{k}$ & \textbf{$\bm{s}$} & \textbf{$\bm{p}$} &  \textbf{$\bm{\dout}$}  & \textbf{$\bm{C}$}  & \textbf{$\bm{\rout}$}            & \textbf{Weights} &  \textbf{Add.} & \textbf{Mul.} & \textbf{Reg.} & \textbf{$2\!:\!1$} & \textbf{MAX} & \textbf{KPU} & \textbf{FCU} & \textbf{PPU}  \\ % & \textbf{$\bm{\rout}$}
\textbf{} & \textbf{(h,w,$\bm{d_{\ell \scalebox{0.75}[1.0]{-}1}}$)} & & & &&&& & \textbf{(h,w,$\bm{d_{\ell \scalebox{0.75}[1.0]{-}1}}$,$\bm{\dout}$)} & \textbf{} & \textbf{} & \textbf{} & \textbf{MUX} & \textbf{} & \textbf{} & \textbf{} & \textbf{}  \\
\midrule
C1 & (24,24,1)  & 24 & 5 & 1 & 2 & 8  &1  &  8      & (5,5,1,8) & 200     & 200     & 800     & 0       & 0       & 8       & 0       & 0         \\
P1 & (24,24,8)  & 24 & 2 & 2 & 0 & 8  &1  &  2      & -         & 0       & 0       & 200     & 0       & 24      & 0       & 0       & 8     \\
C2 & (12,12,8)  & 12 & 5 & 1 & 2 & 16 &4  &  4      & (5,5,8,16)& 816     & 800     & 6.7k    & 2.4k    & 0       & 32      & 0       & 0    \\
P2 & (12,12,16) & 12 & 3 & 3 & 0 & 16 &4  &  4/9    & -         & 0       & 0       & 416     & 108     & 32      & 0       & 0       & 4      \\
F1 & (256)      & 4 & 4  & 4 & - & 10 &320&  $0.02$ & (256,10)  & 8       & 8       & 10      & 2.6k    & 0       & 0       & 2       & 0        \\
\cmidrule(lr){1-18}
\textbf{Sum.} & & \textbf{-} &  \textbf{-}  & \textbf{-} & \textbf{-} & \textbf{-} & \textbf{-} &  \textbf{-} & \textbf{5960}  & \textbf{1024} &  \textbf{1008} & \textbf{8.1k } & \textbf{5.1k } & \textbf{56} & \textbf{40} & \textbf{2} & \textbf{12}  \\
\bottomrule
\end{tabular}
\end{table*}
The table shows the shape of the input and weight tensors, and the stride setting $s$ for each layer. The last dimension of the input tensor represents the number of input channels. The last dimension of the weight tensor represents the number of output channels.
The number of weight configurations $C$ shows that the first two layers are fully parallel in the example with $C=1$. The average number of valid output pixels produced per clock cycle, $\rout$, is also included in the table for each layer.
The table also shows the number of KPUs, PPUs, FCUs, and MAX units per layer.
Also, the number of adders, multipliers, $2\!:\!1$ multiplexers, and registers needed to implement each layer are shown in the table.
To unify the costs of different $N\!:\!1$ multiplexers, we assume that an $N\!:\!1$ multiplexer can be implemented using $N-1$ $2\!:\!1$ multiplexers.
It also has to be noted that the costs do not include the ReLu implementation and the control circuits.
We estimate that those costs are insignificant as control circuits are always on a per-layer basis, and the ReLu activation function is simple and a static scalar function.

The continuous-flow implementation of the running example is shown in Fig.~\ref{fig.running_example}. Components are grouped with curly braces. The interleaving components for a layer are grouped separately and are labeled with an "-IL" suffix.
Each layer in a CNN has $\dout$ output channels by design.
For example, a convolutional layer with 16 filters has 16 output channels.
Each output channel has a non-continuous output when $s>1$.
All output channels together form the output data rate $\rout$ of the layer, producing $\rout$ valid outputs per clock cycle on average.
By interleaving multiple input channels $\din$ together, a continuous flow of data can be ensured. This requires that components can handle multiple configurations to process interleaved input data.
In the following, we will analyze the flow of data in CNNs and then discuss how to interleave data, adapt our components to process interleaved data, and how to build common CNN layers based on the components to ensure a continuous flow of data.

\subsection{Data Rates at the CNN Layers}
\label{sec.data_rates}

Starting from the first layer, the data rate of each layer, $\rout$, is calculated according to (\ref{eq.rout}).
The input data rate for the first layer, $r_0$, is equal to $d_0$, which is the depth of the input image.

\subsection{Interleaving Convolutional Layers}
\label{sec:interleaving_conv}

When the data rate at the output of a layer is reduced due to a stride larger than one, we use pipeline interleaving\cite{p99} to restore continuous flow. Pipeline interleaving is based on the fact that the dataflow has idle times with non valid data. Thus, the valid data from multiple branches is interleaved into a single branch to form a flow of data with less or without idle times.

% PART 2
An example for an interleaved convolutional layer is shown in the running example in Fig.~\ref{fig.running_example} with C2-IL showing the interleaving circuit and C2 showing the interleaved convolutional layer.
The circuit implements a convolutional layer with 16 filters ($\dout = 16$), 8 input channels ($\din = 8$) and an input data rate of $\rin = 2$.
This leads to $8$ input signals that are interleaved and form $2$  output signals that provide a continuous flow of data for the connected KPUs.
Note that each KPU calculates 4 different kernels and each filter consists of 2 KPUs summing up 4 kernel outputs each.

\begin{figure}[h]
    \centering
    \includegraphics{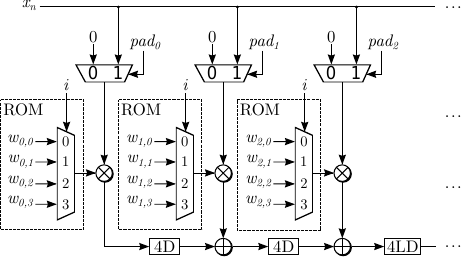}
	\caption{An example of an adapted KPU that can process interleaved data. This KPU processes 4 different $3 \times 3$ kernels out of 4 interleaved channels.}
    \label{fig.KPU1_mul_pad}
\end{figure}

% PART 3
The number of KPUs per convolutional layer thereby depends on the input data rate and the number of filters according to
%The number of KPUs is thereby defined as:
\begin{equation}
\label{eq.kpu_amount}
\text{\#KPUs} = \lceil \rin \rceil \cdot \dout   \,   .
\end{equation}
To calculate all $\din$ kernels for an output channel (filter), each KPU has to calculate $\frac{\din}{ \lceil \rin \rceil}$ different kernels.
This leads to $\din \cdot \dout$ signals to be interleaved into $\lceil \rin \rceil \cdot \dout$ signals that have their continuous-flow restored and can be fed into one KPU each.
An example of a KPU that can calculate 4 different kernels is shown in Fig.~\ref{fig.KPU1_mul_pad}.
Each cycle, the KPU switches to the next kernel and buffers all intermediate results via pipeline interleaving.
The Multiplexers used for the weights can be implemented as a ROM on the FPGA.

\begin{figure}[t]
    \centering
    \includegraphics{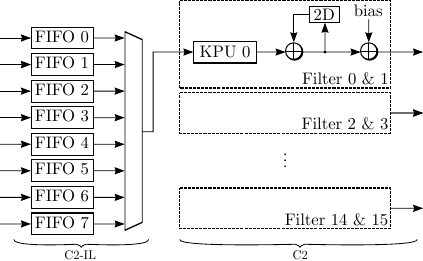}
	\caption{The implementation of a convolutional layer where the input data rate $\rin=0.5$. This allows one KPU to calculate two filters.}
    \label{fig.low_dr_c2}
\end{figure}

In case the data rate $\rout$ is reduced below 1, interleaving can not restore a continuous flow of data, stalling the connected KPUs.
For example, with $\rin = 0.5$, the current input pixel would be present for two clock cycles and the KPU can calculate the corresponding kernel of two different filters, one after the other, by reconfiguring the KPU.
The number of configurations per KPU is then calculated as
\begin{equation}
%C = \min \left (\left \lceil \frac{\din}{2^{ \lceil \text{log}_2( \rin ) \rceil }} \right \rceil, \din \dout \right).
\label{eq.kpu_configs_amount}
C = \min \left (\left \lceil \frac{\din}{\rin} \right \rceil, \din \dout \right).
\end{equation}
An example is shown in Fig.~\ref{fig.low_dr_c2} where a convolutional layer is implemented with $\rin = 0.5$, $\din=8$, and $\dout=16$.
Each KPU now calculates 8 kernels for two filters, leading to 16 configurations for the KPU.
With each pixel present for two clock cycles, the KPU processes the current input pixel and calculates the current kernel of the two filters sequentially.
The summation has to be interleaved to calculate two separate sums for the two filters.
The number of interleaved channels $I$ is then calculated as
\begin{equation}
\label{eq.interleaving_amount}
I = \left \lceil \frac{C}{\din} \right \rceil.
\end{equation}
This adaption changes the number of KPUs to
\begin{equation}
\label{eq.kpu_amount_better}
\text{\#KPUs} = \lceil \rin \rceil \cdot \frac{\dout}{I} \, ,
\end{equation}
and allows for a more efficient implementation that reduces idle times even more by allowing the architecture to use even fewer components per layer when the data rate is low.
This is especially important for the depthwise-separable convolution, which consists of fewer kernels compared to the normal convolution.
This type of convolutional layer consists of a depthwise convolution followed by a pointwise convolution.

\begin{figure}[t]
    \centering
    \includegraphics{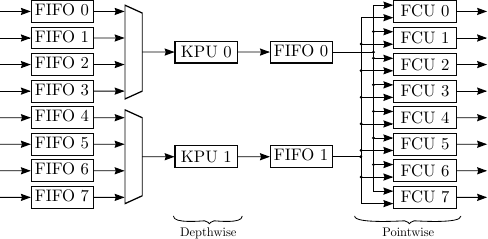}
	\caption{An example for an interleaved depthwise-separable convolution with $\din=8$ input channels and a data rate of $\rin = 2$.}
    \label{fig.dps_conv}
\end{figure}

The depthwise convolution has $g$ groups, where $g$ is a multiple of $\din$.
For simplification, we assume the most prominent setting for $g$ with $g=\din$.
So each output channel (group) processes a single input channel with a single kernel.
Compared to the standard convolution, each channel in the depthwise convolution only depends on a single input channel.
%This allows interleaving and saving components similar to the standard convolution shown before.
This removes the $\dout$ factor when calculating the number of KPUs needed for a depthwise convolution, leading to
\begin{equation}
\label{eq.kpu_dw_amount}
\text{\#KPUs} =  \lceil \rin \rceil.
\end{equation}
Without $\dout$ as a factor, the number of configurations per KPU is then calculated as
\begin{equation}
%C = \min \left (\left \lceil \frac{\din}{2^{ \lceil \text{log}_2( \rin ) \rceil }} \right \rceil, \din \right).
\label{eq.kpu_configs_dw_amount}
C = \min \left (\left \lceil \frac{\din}{\rin} \right \rceil, \din \right).
\end{equation}
The pointwise convolution is a standard convolution with a $1 \times 1$ kernel.
With an $1 \times 1$ KPU consisting of just a single multiplier, the output of a channel would consist of $\din$ multiplications summed up over $\frac{\din}{\rin}$ clock cycles.
The pointwise convolution can thereby be implemented as a fully connected layer, where each FCU is configured following (\ref{eq.biggest_fcu}).
An example of a depthwise-separable convolution is shown in Fig.~\ref{fig.dps_conv}.
The input data rate is $\rin = 2$ with $\din = 8$ input channels. The output data rate of the depthwise convolution stays at $\rout = 2$ and is fed into the pointwise convolution components. Each FCU processes $j=\jmax=2$ input pixels and calculates $h=\hmax=1$ output channels because $\rin = \frac{\jmax}{\hmax}=\frac{2}{1}$. Thereby, only 8 FCUs are needed.

\subsection{Interleaving Pooling Layers}
\label{sec:interleaving_pooling}
\begin{figure}[t]
    \centering
    \includegraphics{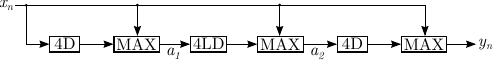}
	\caption{An example for an interleaved $2 \times 2$ max-pooling operation with 4 channels.}
    \label{fig.2x2max_pool_mul}
\end{figure}

Interleaving pooling layers works similarly to convolutional layers, except that each output channel only processes one input channel.
This leads to $\din$ input signals to interleave.
Note also that pooling layers do not change the number of channels, i.e., $\din = \dout$. With each PPU processing one pixel per clock cycle, the equation for the number of PPUs breaks down to
\label{eq.ppu_amount}
\begin{equation}
\text{\#PPUs} = \lceil \rin \rceil.
\end{equation}
An example for an interleaved PPU is shown in Fig.~\ref{fig.2x2max_pool_mul}. The circuit implements a $2 \times 2$ max-pooling operation for 4 interleaved channels using pipeline interleaving.

\subsection{Interleaving Fully Connected Layers}
\label{sec:interleaving_fc}
The FCU circuit in Fig.~\ref{fig.fc} produces a continuous flow at the output despite a non-continuous flow of data at the input because the FCU generates valid outputs for each neuron sequentially with the last batch of input values.
Therefore, a fully connected layer does not need multiplexers or interleaving of the input.
Each FCU simply processes multiple input features at once and calculates multiple output channels sequentially.

\section{Complexity analysis}
\label{sec:Comparison}

In the following, we will discuss the complexity of the KPU, PPU, and FCU components, how the complexity of layers changes regarding  the input data rate, and how our approach compares to a fully parallel implementation of common CNN models.

\subsection{Complexity of Data Interleaving}

The cost of data interleaving is split into the cost of registers and multiplexers.
%The cost for multiplexers and registers is determined by $I$, $C$, $\dout$, and $\rout$ of the previous layer.
Regarding multiplexers, $\frac{\dout}{I}$ signals are interleaved into $\rout$ signals to ensure a continuous flow of data. This leads to

\begin{equation}
    \label{eq.cost_IL_mux}
    \text{\#MUX} = \lceil \rin \rceil \left ( \frac{\din}{\lceil \rin \rceil I} - 1 \right )
    = \frac{\din}{I} - \lceil \rin \rceil  \\
\end{equation}

$2\!:\!1$ multiplexers where each $N\!:\!1$ multiplexer is implemented using $N-1$ $2\!:\!1$ multiplexers.
The previous layer will produce $\dout$ valid outputs every $C$ clock cycles. Those valid outputs are spread over the last $I$ of $C$ clock cycles. Therefore,

\begin{equation}
    \label{eq.cost_IL_fifos}
    \text{\#Registers} = \frac{\dout}{I} I = \dout \\
\end{equation}

registers are needed for the FIFOs.

\subsection{Complexity of the KPU}

The parameters of a KPU component consist of the kernel size, $k$, the feature map size, $f$, and the number of configurations $C$.
The number of adders and multipliers depends on the kernel size $k$ because there are $k^2$ multiplications in the transposed form that have to be accumulated. This leads to
\begin{equation}
    \label{eq.cost_kpu_adders}
    \text{\#Adders} = k^2-1 \, , \\
\end{equation}
\begin{equation}
    \label{eq.cost_kpu_mult}
    \text{\#Multipliers} = k^2 \, . \\
\end{equation}
To store all partial sums, each KPU has $k(k-1)$ registers and $k-1$ line buffers.
Each line buffer consists of $f-k+1$ registers. The KPU has to store all partial sums for each configuration. This leads to a number of registers for each KPU equal to
\begin{equation}
    \label{eq.cost_kpu_reg}
    \text{\#Registers}\! =\! \left ( k(k\!-\!1)\! + \!(k\!-\!1)(f\!-\!k\!+\!1) \right ) C  \, . \\
\end{equation}
The number of multiplexers also depends on the number of configurations and the kernel size. For all $k^2$ multipliers, $C-1$ $2\!:\!1$ multiplexers are needed to switch between the weights of the configurations, leading to a number of $2\!:\!1$ multiplexers equal to
\begin{equation}
    \label{eq.cost_kpu_mux}
    \text{\#MUX} = k^2(C-1)   \, . \\
\end{equation}

\subsection{Complexity of the Convolutional Layer Channel Accumulation}
Each output channel (filter) in a convolutional layer processes all input channels.
By removing the multipliers, the FCU can be repurposed for the accumulation.
For each output channel, $j=\lceil \frac{\text{\#KPUs}}{\dout} \rceil$ inputs are accumulated per clock cycle and produce $h=\lceil \frac{C}{\din} \rceil = I$ interleaved outputs over $C$ clock cycles.
Each output signal of a convolutional layer that interleaves $I$ output channels needs one accumulator component. This leads to $\frac{\dout}{I}$ accumulators per convolutional layer. The cost for the accumulation is then defined by:
\begin{equation}
    \label{eq.cost_conv_accu_reg}
    \text{\#Registers} = \frac{\dout}{I} I = \dout \, , \\
\end{equation}
\begin{equation}
    \label{eq.cost_conv_accu_adders}
    \text{\#Adders} = \frac{\dout}{I} \left \lceil \frac{\text{\#KPUs}}{\dout} \right \rceil  \, . \\
\end{equation}
There is a special case when $\din=1$ and $\rin=1$ where no accumulation is needed. This is shown in the running example at (C1). In this case, with $\din=1$, the sum consists of a single value. Thereby, no accumulation is needed.

\subsection{Complexity of the Bias}
Some layers add an individual bias to each output channel. This leads to
\begin{equation}
    \label{eq.cost_bias_adders}
    \text{\#Adders} = \frac{\dout}{I} \\
\end{equation}
adders for each interleaved output. Each adder has an $I$:1 multiplexer. This adds a multiplexing cost of
\begin{equation}
    \label{eq.cost_bias_mux}
    \text{\#MUX} = \frac{\dout}{I} (I - 1) = \dout - \frac{\dout}{I} \\
\end{equation}
$2\!:\!1$ multiplexers in total.

\subsection{Complexity of the PPU}

%The complexity of the PPU breaks down into the number of MAX units
The PPU shares similar complexity characteristics to a KPU with
\begin{equation}
    \label{eq.cost_ppu_max}
    \text{\#MAX} = k^2-1 \, , \\
\end{equation}

and the same complexity for registers shown in (\ref{eq.cost_kpu_reg}).

\subsection{Complexity of the FCU}

The FCU component has $j$ inputs and calculates $h$ different sums of products. Each clock cycle, the weights are switched and a different weighted sum of all $j$ inputs is calculated until all $\din$ inputs are processed. This results in
\begin{equation}
    \label{eq.cost_fcu_mul}
    \text{\#Multipliers} = j \, , \\
\end{equation}
\begin{equation}
    \label{eq.cost_fcu_mux}
    \text{\#MUX} = j(C-1) \, . \\
\end{equation}
The weighted sum is accumulated using one adder and one register, leading to
\begin{equation}
    \label{eq.cost_fcu_adders}
    \text{\#Adders} = j \, , \\
\end{equation}
\begin{equation}
    \label{eq.cost_fcu_regs}
    \text{\#Registers} = h \, .\\
\end{equation}

\subsection{Impact of the Data Rate on the Complexity}
By interleaving data and introducing reconfigurable components to handle interleaved data, the number of components is reduced, and the utilization of the components is high.
Next, we discuss these effects based on the convolutional layer, which is the most prominent layer in a CNN.

Table~\ref{td.conv_layer_costs}  shows  the resources for the implementation of a convolutional layer with different data rates.
The convolutional layer has the settings $f=28$, $k=7$, $p=3$, $\din=8$, $\dout=16$, and the input data is not interleaved.
The costs for FIFOs and data interleaving are left out because they depend on the previous layer.
The first row of the table shows the resources for a fully parallel implementation where the data rate is equal to $\din$.
For each row below, the data rate is halved.
It can be seen that the number of KPUs is proportional to the data rate, and thereby is halved for each row.
By halving the number of KPUs, the number of adders and multipliers also halves.
But, each KPU has to implement more and more configurations because data is interleaved more intensely to achieve a continuous flow with the decreasing data rate.
This results in higher multiplexer costs.
The number of registers stays the same, but they are organized differently based on the data interleaving.
\begin{table}[t]
\centering
\caption{A resource comparison of the implementation of a convolutional layer with different input data rates.}
\label{td.conv_layer_costs}
\begin{tabular}{cccccc}
\toprule
$\bm{\rin}$ & \textbf{Add.} & \textbf{Mul.} & \textbf{Reg.} & \textbf{$2\!:\!1$ MUX} & \textbf{KPUs}   \\
\cmidrule(lr){0-0} \cmidrule(lr){2-6}
8        & 6,272     & 6,272     & 22,288    & 0        & 128     \\
4        & 3,136     & 3,136     & 22,288    & 3,136     & 64      \\
2        & 1,568     & 1,568     & 22,288    & 4,704     & 32      \\
1        & 784      & 784      & 22,288    & 5,488     & 16      \\
0.5      & 392      & 392      & 22,288    & 5,880     & 8       \\
0.25     & 196      & 196      & 22,288    & 6,076     & 4       \\
0.125    & 98       & 98       & 22,288    & 6,174     & 2       \\
0.0625   & 49       & 49       & 22,288    & 6,223     & 1       \\
0.03125* & 49       & 49       & 22,288    & 6,223      & 1       \\
\bottomrule
\end{tabular}
\\
\vspace{1mm}
\footnotesize{*The input data rate leads to a stall.}
\end{table}
The last row shows an example where the data rate is so low that a continuous flow of data cannot be restored by interleaving. Thereby, the KPUs will stall.
For the depthwise-separable convolution, the same observations can be made regarding the number of adders, multipliers, and registers. The resource comparison for the depthwise-separable convolution is shown in Table~\ref{td.conv_dw_layer_costs}.
For this table, the same parameters were used as in Table~\ref{td.conv_layer_costs}.
\begin{table}[t]
\centering
\caption{A resource comparison of the implementation of a depthwise-separable convolutional layer with different input data rates.}
\label{td.conv_dw_layer_costs}
\begin{tabular}{ccccccc}
\toprule
$\bm{\rin}$ & \textbf{Add.} & \textbf{Mul.} & \textbf{Reg.} & \textbf{$2\!:\!1$ MUX} & \textbf{KPUs} & \textbf{FCUs}   \\
\cmidrule(lr){0-0} \cmidrule(lr){2-7}
8        & 512      & 520      & 1,416     & 0       & 8        & 16       \\
4        & 256      & 260      & 1,416     & 260     & 4        & 16       \\
2        & 128      & 130      & 1,416     & 390     & 2        & 16       \\
1        & 64       & 65       & 1,416     & 455     & 1        & 16       \\
0.5*      & 56       & 57      & 1,416     &463      & 1        & 8        \\
0.25*     & 52       & 53      & 1,416     &467      & 1        & 4        \\
\bottomrule
\end{tabular}
\\
\vspace{1mm}
\footnotesize{*The input data rate leads to a stall.}
\end{table}

\begin{table}[t]
\centering
\caption{A comparison between a fully parallel implementation and our continuous-flow approach for different models.}
\label{td.model_compare} % l<{\hspace{0mm}}
\begin{tabular}{@{\hspace{1mm}}C{12.02mm}@{\hspace{2mm}}C{8.8mm}   C{5.55mm}@{\hspace{2mm}}   C{7.25mm}@{\hspace{2mm}}C{7.25mm}@{\hspace{2mm}}C{5.45mm}@{\hspace{2mm}} C{7.25mm}@{\hspace{2mm}}C{6.9mm}@{\hspace{2mm}}C{6.7mm}@{\hspace{1mm}}}
\toprule
\textbf{Model} & \textbf{Param.} & \textbf{Imp.} & \textbf{Add.} & \textbf{Mul.} & \textbf{Reg.} & \textbf{$2\!:\!1$ MUX} & \textbf{KPUs} & \textbf{FCUs}  \\
\midrule
Running & \multirow{2}{*}{6.0k    } & Ref.      & 6.0k    & 6.0k    & 8.1k    & 0          & 136     & 10           \\
example         &          & Ours     & 1.0k    & 1.0k    & 8.1k    & 5.1k        & 40      & 2            \\
\cmidrule(lr){3-9}
MobileNet  & \multirow{2}{*}{470k  } & Ref.  &   475k  & 476k  & 76k   &0         & 1.5k    & 2.5k       \\
      $\alpha$=0.25   &          & Ours     & 1.1k    & 1.1k    & 76k   & 477k        & 44      & 632            \\
\cmidrule(lr){3-9}
MobileNet & \multirow{2}{*}{1.3M    } & Ref.      & 1.3M    & 1.3M    & 151k  &0          & 3.0k    & 4.0k     \\
     $\alpha$=0.5    &          & Ours     & 3.4k    & 3.5k    & 151k  & 1.3M         & 80      & 2.2k         \\
\cmidrule(lr){3-9}
MobileNet & \multirow{2}{*}{2.6M    } & Ref.      & 2.6M    & 2.6M    & 226k  &0          & 4.6k    & 5.5k     \\
    $\alpha$=0.75     &          & Ours     & 7.2k    & 7.2k    & 249k  & 2.6M         & 122     & 1.9k          \\
\cmidrule(lr){3-9}
MobileNet  & \multirow{2}{*}{4.2M    } & Ref.      & 4.3M    & 4.3M    & 300k  &0        & 6.1k    & 7.0k    \\
    $\alpha$=1.0     &          & Ours     & 12.2k   & 12.2k   & 300k  & 4.3M       & 158     & 5.5k          \\
\cmidrule(lr){3-9}
\multirow{2}{*}{ResNet18} & \multirow{2}{*}{11.7M   } & Ref.   &   11.7M   & 11.7M   & 30M   &0      & 1.2M    & 1.9k           \\
         &          & Ours     & 33.7k   & 33.7k   & 30M   & 11.7M      & 2.8k    & 514          \\
\bottomrule
\end{tabular}
\end{table}
It can be seen that the depthwise-separable convolution needs far fewer resources than the normal convolution.
Our approach still achieves a significant reduction in adders and multipliers for high data rates where $\rin \geq 1$. Not only is the number of KPUs reduced, but also the number of inputs for each FCU. Both reductions have a significant impact on the number of adders and multipliers. In the low data rate regime ($\rin < 1$), the number of KPUs has already dropped to 1, and only the number of FCUs can be reduced. This results in a marginal reduction of resources. Also, the KPU is stalling because the continuous flow cannot be restored anymore.

\section{Implementation of Common CNN models}
\label{sec:CommonCNNs}
To evaluate our approach, we have implemented\footnote{ \text{https://github.com/numericsgithub/CNN-Flow-Paper}} a MobileNetV1 \cite{mobilenet} model and analyzed the ResNet18 from the ResNet model family\cite{resnet}.
All models were trained on the ImageNet \cite{ImageNet} dataset.
The MobileNetV1 architecture heavily relies on the depthwise-separable convolution and consists of one convolutional layer, 13 depthwise-separable convolutional layers, followed by an average pooling layer and a fully connected layer.
The number of filters in the convolutional layers is controlled by the factor $\alpha$.
The MobileNetV1 model family consists of 4 variations with $\alpha \in \{0.25, 0.5, 0.75, 1.0\}$.
The ResNet18 architecture consists of residual layer blocks that are put in sequence, where the
suffix 18 in the name stands for the number of layers in the model.
In Table~\ref{td.model_compare}, we compare 
the number of parameters and the resource costs for our proposed architecture (Ours) and for a fully parallel implementation (Ref.).
Both implementations always have the same input data rate.
The costs for the implementation of the activation functions are not included.
Activation functions are stateless, scalar functions that do not have to be adapted to a continuous flow architecture, and thereby are not considered in the comparison.
The costs for the layer control circuits, for example, the costs for counters to keep track of the current row and column of the input pixel, are also not included as they only apply on a per-layer basis.

The Mobilenet and ResNet models use a single average pooling layer before the fully connected layer.
This can be implemented using a depthwise convolutional layer with constant weights.
For example, the average of four pixels can be calculated with a $2 \times 2$ KPU where all weights are set to $1/4$.
Another detail is the implementation of residual layers. Considering the residual layers in the ResNet model family, the merging of two layers is implemented by addition.
The output pixel of the residual layer is calculated when both the pixel of the shortcut and the pixel of the sequential part of the block are calculated.
We thereby assume that the layer after the merged activations has an input data rate equal to the lowest data rate of the two merged layers.

By analyzing the table, it can be observed that the number of adders and multipliers in the proposed approach is reduced substantially compared to the fully parallel implementation.
The running example needs around $1/6$ of adders and multipliers, whereas the MobileNet and ResNet models reduce the number of adders and multipliers by several orders of magnitude.
By contrast, reconfigurable components, multiplexing between weights and input channels is needed in the proposed approach.
%For example, the MobileNet $\alpha$=0.25 model needs 477k multiplexers when implemented with the proposed approach.
%But, this reduces the number of adders and multipliers from 475k down to 1.1k, and almost
However, almost all multiplexers can be implemented using BRAM, because multiplexing between weights represents a read-only memory and the multiplexing cost of interleaving input channels is low as the data rate decreases.
Another aspect to consider is that the number of registers does not change when our continuous-flow approach is applied, except for the MobileNet $\alpha$=0.75.
This specific model has filter sizes that are not base two.
This leads to a rounding in (\ref{eq.interleaving_amount}), rounding up the number of KPUs needed.
This breaks the continuous flow and adds register costs.
However, the increase in registers is small compared to the massive savings in adders and multipliers.

\section{Synthesis Results}
\label{sec:Synthesis}

The first experiment, we implemented the MobileNetV1, synthesized it using Vivado and compared it to recent state-of-the-art FPGA implementations.
Table~\ref{tab:mobilenet_impl_metadata} shows the results.
To implement the CNN,  we performed a quantization-aware training to adapt the model to the 8-bit fixed-point format used in the implementation.
For the quantization-aware training, we have used the MQUAT framework\footnote{MQUAT: \MQUATLink}. Using MQUAT, the model has been trained to use 8-bit wide fixed-point formats for weights and activations. Only the output activations of the last layer have been set to 12 bits. The process to implement CNNs has been fully automated by developing a code generator.
To implement all arithmetic units, FloPoCo \cite{de_dinechin2024application} has been used except for the general multipliers, where we used the same concept shown in \cite{main_ref}, where a single DSP implements two multiplications.
The code generator also provides a bit- and clock-accurate simulator for precise accuracy determination using TensorFlow to allow fast inference under hardware conditions.
The code generator performs a worst-case analysis to calculate the width of the adders in the KPUs and FCUs of the design. With the given equations, the code generator can implement layer after layer by calculating the data rate for the next layer based on (\ref{eq.rout}), which then allows the software to generate the right number of FCUs/KPUs needed.
Thereby, the code generator derives the number of units, configurations, etc., automatically.
\begin{table}[t]
    \centering
    
    \caption{Comparison of MobileNetV1 implementations}
    \label{tab:mobilenet_impl_metadata}
    \begin{tabular}{lllll}
    \toprule
         & FINN \textbf{\cite{finn_r}}* & \textbf{\cite{main_ref}} & \textbf{\cite{dcnn_acc}} & \textbf{Ours} \\
        \midrule
        Freq. (MHz)      & 333              & 211         & 250         & \textbf{350} \\
        LUT Util         & 501,363          & 412,354     & 402,200     & \textbf{204,931} \\
        FF Util          & \textbf{476,316} & 991,909     & -           & 563,255 \\
        DSP Util         & \textbf{106}     & 5,852       & 6,414       & 5,691 \\
        BRAM Util        & 898              & 1,838.5     & \textbf{214}& 1,702.5 \\
        URAM Util        & 0                & 0           & 394         & 0 \\
        %Multipliers     &                  &  11336      &             & 12,198 \\
        FPGA             & Alveo U280       & XCVU37P     & XCVU9P      & XCVU37P \\
        Power (W)        & 41.69            & 39.465      & -           & 24.646\\
        FPS (Inf./s)     & 925              & 4,205.5     & 2,637       & \textbf{6,944.44} \\
        E. Eff. (mJ/Inf.)& 45.07            & 9.38        & -           & \textbf{3.55}\\
        Latency (ms)     & -                & 0.60        & 379.21      & \textbf{0.37} \\
        Precision        & 4-bit            & 8-bit       & 8-bit       & 8-bit             \\
        Top-1 acc.       & 70.4 \%          &  70.1\%     & -           & \textbf{70.5\%} \\
        %fp32 Top-1 acc.  &                 &  70.9\%    &             & \textbf{70.9\%} \\
        \bottomrule
    \end{tabular}
    \\
    \vspace{1mm}
    \footnotesize{*FINN implementation results are from \textbf{\cite{lutmul_ref}}}
\end{table}

\begin{figure}[t]
    \centering
    \includegraphics[width=1.0\linewidth]{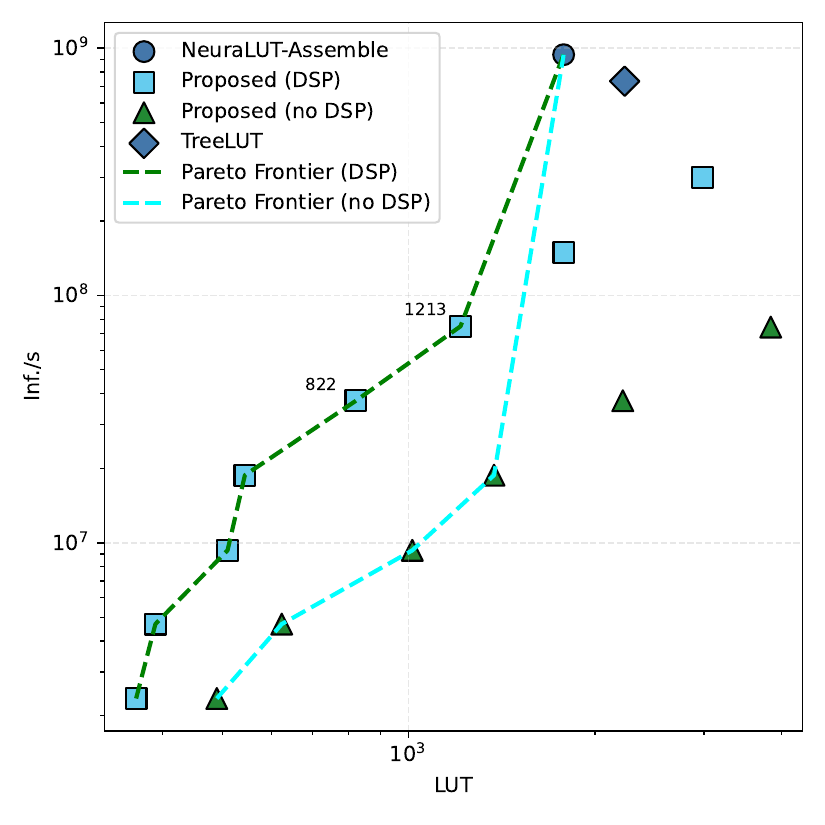}
	\caption{Throughput compared to LUT utilization for our approach compared to the state-of-the-art parallel implementations.}
    \label{fig.jsc_acc}
\end{figure}

\begin{table*}[t]
\centering
\centering
\caption{Comparison with fully parallel architectures on the JSC dataset.}
\label{td.jsc_compare}

\begin{tabular}{cccccccccc}
\toprule
\textbf{} & \textbf{Accuracy} & \textbf{$\bm{r_{0}}$} & \textbf{$\bm{\text{F}_{\text{max}}}$} & \textbf{LUT} & \textbf{FF} & \textbf{BRAM} & \textbf{DSP} & \textbf{Speed} & \textbf{latency}  \\
                    &          &          & MHz      &          &          &          &          & MInf/s   & ns        \\
\midrule
PolyLUT (JSC-XL) \cite{lou2024polylut}   & 75\%     & 16       & 235      & 236,541  & 2,775    & 0        & 0        & 235.0    & 21.0      \\
NeuraLUT (JSC-5L) \cite{neural_LUT}  & 75\%     & 16       & 368      & 92,357   & 4,885    & 0        & 0        & 368.0    & 14.0      \\
NeuraLUT-Assemble \cite{andronic2025neuralut-assemble}  & 76.0\%   & 16       & 941      & 1,780    & 540      & 0        & 0        & 941      & 2.1       \\
TreeLUT  \cite{TreeLUT}           & 75.6\%   & 16       & 735      & 2,234    & 347      & 0        & 0        & 735      & 2.7       \\
DWN   \cite{DWN}                    & 76.3\%   & 16       & 695      & 6,302    & 4,128    & 0        & 0        & 695      & 14.4      \\
hls4ml  \cite{hls4ml }                  & 76.2\%   & 16       & 200      & 63,251   & 4,394    & 0        & 38       & 200      & 45.0      \\
\cmidrule(rl){1-10}
Proposed (DSP)      & 75.2\%   & 16       & 690      & 5,308    & 19,162   & 0.0      & 184      & 690.1    & 72.5      \\
Proposed (DSP)      & 75.2\%   & 8        & 600      & 2,984    & 12,215   & 3.5      & 149      & 300.0    & 93.3      \\
Proposed (DSP)      & 75.2\%   & 4        & 596      & 1,782    & 7,048    & 3.5      & 76       & 149.0    & 104.0     \\
Proposed (DSP)      & 75.2\%   & 2        & 600      & 1,213    & 4,450    & 3.5      & 38       & 75.0     & 120.0     \\
Proposed (DSP)      & 75.2\%   & 1        & 600      & 822      & 2,535    & 3.5      & 35       & 37.5     & 160.0     \\
Proposed (DSP)      & 75.2\%   & 1/2      & 600      & 544      & 1,610    & 4.0      & 19       & 18.8     & 240.0     \\
Proposed (DSP)      & 75.2\%   & 1/4      & 596      & 510      & 1,230    & 2.0      & 11       & 9.3      & 339.0     \\
Proposed (DSP)      & 75.2\%   & 1/8      & 600      & 390      & 765      & 2.5      & 5        & 4.7      & 655.0     \\
Proposed (DSP)      & 75.2\%   & 1/16     & 600      & 363      & 639      & 2.5      & 3        & 2.3      & 1,060.0   \\
Proposed (no DSP)   & 75.2\%   & 16       & 564      & 12,857   & 23,728   & 0.0      & 0        & 564.0    & 86.9      \\
Proposed (no DSP)   & 75.2\%   & 8        & 564      & 10,513   & 16,503   & 3.5      & 0        & 282.0    & 97.5      \\
Proposed (no DSP)   & 75.2\%   & 4        & 556      & 6,529    & 10,287   & 3.5      & 0        & 139.0    & 109.7     \\
Proposed (no DSP)   & 75.2\%   & 2        & 596      & 3,846    & 6,350    & 3.5      & 0        & 74.5     & 119.1     \\
Proposed (no DSP)   & 75.2\%   & 1        & 600      & 2,218    & 4,007    & 3.5      & 0        & 37.5     & 158.3     \\
Proposed (no DSP)   & 75.2\%   & 1/2      & 600      & 1,374    & 2,474    & 4.0      & 0        & 18.8     & 238.3     \\
Proposed (no DSP)   & 75.2\%   & 1/4      & 596      & 1,014    & 1,775    & 2.0      & 0        & 9.3      & 337.3     \\
Proposed (no DSP)   & 75.2\%   & 1/8      & 600      & 624      & 1,027    & 2.5      & 0        & 4.7      & 653.3     \\
Proposed (no DSP)   & 75.2\%   & 1/16     & 600      & 490      & 781      & 2.5      & 0        & 2.3      & 1,058.3   \\

\bottomrule
\end{tabular}
\end{table*}

All power results in Table~\ref{tab:mobilenet_impl_metadata} are obtained using the Vivado power analysis, with accurate switching activity obtained from a post-synthesis simulation.
In general, our approach reaches the highest number of FPS with the lowest latency and the best energy efficiency with just 3.55 mJ per inference, whereas the FINN \cite{finn_r} implementation shows a much more efficient use of DSP resources relative to the FPS. Compared to \cite{main_ref} on the same FPGA, the number of DSPs and BRAMs used in the proposed CNN is lower by a small margin.
The number of LUTs and registers is almost cut in half.
Furthermore, our implementation allows a far higher frequency, which reduces the latency by almost half and almost doubles the throughput. Finally, with the quantization-aware training, we reach a better \mbox{top-1} accuracy for our implementation.

In the second experiment, we performed an analysis for a wide variety of implementations of different data rates on a well known jet substructure tagging dataset JSC \cite{jsc_dataset}, which was used in several previous FPGA-based implementations \cite{lou2024polylut,neural_LUT,andronic2025neuralut-assemble,TreeLUT,DWN,hls4ml}.
These implementations were all fully parallel  (i.e., one hardware unit per neuron) and all of them except hls4ml are LUT-based, i.e., do no use any DSP (and can not use them).
Our DNN consists of two dense layers, which have 16 neurons each, with a final third dense layer with 5 neurons.
This DNN was then trained with MQUAT,
%on the jet substructure tagging dataset JSC \cite{jsc_dataset}, 
reaching $75.2\%$ \mbox{top-1} validation accuracy.

To make our results comparable to the state-of-the-art, we also created implementations that use LUT resources for the multiplications instead. 
The LUT multipliers were generated using the \texttt{IntMultiplier} operator from FloPoCo \cite{flopoco}, which implements the state-of-the-art approach shown in \cite{10631128, 10579405}.
For the synthesis results, we used the Xilinx Virtex Ultrascale+ FPGA (\texttt{xcvu9p-flgb2104-2-i}), as it is the same one used in \cite{amigolut}, \cite{andronic2024neuralut}, and \cite{lou2024polylut}.

The detailed results are given in Table~\ref{td.jsc_compare} while Fig.~\ref{fig.jsc_acc} shows a Pareto plot that compares the tradeoff between throughput and LUT utilization.
The data rate $r_0$ of the previous work (top of Table~\ref{td.jsc_compare}) is identical to the number of inputs (16) while our results can be tuned to different data rates.
The Pareto plot contains all implementations that reached a validation accuracy of at least $75\%$ and use less than 5k LUTs.
We also limit the maximum frequency to the maximum frequency of the FPGA clock tree, which is 800 MHz, as higher frequencies are not possible to reach in practice.

The results shows how flexible our approach is to reach a certain throughput or resource utilization target, and how the Pareto frontier is extended for lower throughput targets.
%As our approach is also not restricted to only use LUT resources but allows to utilize DSPs and BRAM resources too.
The results also show that fully parallel implementations become more efficient at the highest rate $r_0=16$.
This is expected, as our approach takes advantage of the reconfiguration of arithmetic units that become less reconfigurable with higher data rates.

In detail, one can observe that NeuraLUT-Assemble~\cite{andronic2025neuralut-assemble} and TreeLUT \cite{TreeLUT} achieve lowest resource costs at a higher throughput for $r_0=16$.
This is due to the architecture design that is specialized for fully parallel implementations and due to the fact that those architectures do not rely on conventual arithmetic but learn LUT interconnections and truth table contents when training.
This allows for highly resource efficient implementations but does not allow a sequential implementation.
Our approach, in contrast, allows for highly reconfigurable implementations that allows significant lower cost at lower data rates. 
%This is shown in the table by presenting a wide variety of implementations that support different data rates. 
This flexibility enables designers to trade off resource utilization against throughput according to application requirements.
Additionally, as our approach uses conventional arithmetic, it allows utilizing DSP resources too as we can freely decide how to implement multiplications and additions where other approaches that do use LUT-based layers cannot use DSP resources.
Using DSPs, we reach a lower LUT resource utilization than NeuralLUT-Assemble at $r_0=2$, without DSPs it is achieved at $r_0=1/2$.
Using DSPs, we reach a lower LUT resource utilization than NeuralLUT-Assemble at $r_0=2$, without DSPs it is achieved at $r_0=1/2$.
Comparing our approach to state-of-the-art conventional arithmetic designs like hls4ml\cite{hls4ml}, it can be seen that our approach is superior in resources and throughput.

\section{Conclusions}
\label{sec:Conclusions}

In this work, we have proposed a novel approach to design continuous-flow CNN architectures. It is based on analyzing the data flow of the CNN, interleaving data based on the data rate for each layer, constructing reconfigurable components to handle the interleaved data, and thereby ensuring a high utilization close to 100\%. 
Experimental results show that this approach saves a significant number of resources for common CNN models compared to the state-of-the-art and the 1:1 mapping of neurons to hardware units as used in most previous work using unrolled architectures.
Furthermore, the experimental synthesis results demonstrate that our approach also translates to significantly lower resources compared to the state-of-the-art.

\bibliographystyle{IEEEtran}
\bibliography{bibliography}

@STRING{ACM = {Proc. ACM/SIGDA Int. Symp. FPGAs}}

@STRING{APCCAS = {Proc. {IEEE} Asia-Pacific Conf. Circuits Syst.}}

@STRING{ARITH = {Proc. {IEEE} Symp. Comput. Arithmetic}}

@STRING{ASAP = {Proc. {IEEE} Int. Conf. Application-Specific Syst.}}

@STRING{DSP = {Digital Signal Process.}}

@STRING{FPL = {Proc. Int. Conf. Field-Programmable Logic Appl.}}

@STRING{ICASSP = {Proc. {IEEE} Int. Conf. Acoust. Speech Signal Process.}}

@STRING{ICFPT = {Proc. Int. Field Programmable Tech.}}

@STRING{C_ACM = {Comm. of the ACM}}

@STRING{ICCSP = {Int. Conf. Comm. Signal Process.}}

@STRING{ICCV = {IEEE/CVF Int. Conf. Comput. Vis.}}

@STRING{CVPR = {IEEE/CVF Conf. Comput. Vis. Pattern Recognit.}}

@STRING{AIBT = {IEEE Int. Conf. Artificial Intell. Blockchain Internet Things}}

@STRING{FCCM = {IEEE Int. Symp. Field-Programmable Custom Comput. Machines}}

@STRING{EURASIP_JORNAL_SIG = {EURASIP J. Adv. Signal Process.}}

@STRING{JINST = {J. Instrum.}}

@STRING{EACL = {Conf. European Chapter Comm. Comput. Ling.}}

@STRING{AERO = {IEEE Aerosp. Conf.}}

@STRING{ICPP = {Int. Conf. Parallel Process.}}

@STRING{TRETS = {ACM Trans. Reconf. Tech. Syst.}}

@Article{ji2023fpqnet,
    TITLE = {{FPQNet}: Fully Pipelined and Quantized {CNN} for Ultra-Low Latency Image Classification on {FPGAs} Using {OpenCAPI}},
    AUTHOR = {Ji, Mengfei and Al-Ars, Zaid and Hofstee, Peter and Chang, Yuchun and Zhang, Baolin},
    JOURNAL = {Electronics},
    YEAR = {2023},
    MONTH = oct,
    NUMBER = {19},
    ARTICLE-NUMBER = {4085},
    ISSN = {2079-9292},
    DOI = {10.3390/electronics12194085},
    VOLUME = {12}
}

@article{tridgell2019unrolling,
    title = {Unrolling Ternary Neural Networks},
    author = {Tridgell, Stephen and Kumm, Martin and Hardieck, Martin and Boland, David and Moss, Duncan and Zipf, Peter and Leong, Philip H. W.},
    journal = TRETS,
    year = {2019},
    month = Dec,
    volume = {12},
    number = {4},
    issn = {1936-7406},
    doi = {10.1145/3359983},
    articleno = {22},
    numpages = {23}
}

@INPROCEEDINGS{umuroglu2020logicnets,
    title={{LogicNets}: Co-Designed Neural Networks and Circuits for Extreme-Throughput Applications}, 
    author={Umuroglu, Yaman and Akhauri, Yash and Fraser, Nicholas James and Blott, Michaela},
    booktitle=FPL, 
    year={2020},
    month=aug,
    pages={291-297},
    doi={10.1109/FPL50879.2020.00055}
}

@INPROCEEDINGS{wang2019lutnet,
    title={{LUTNet}: Rethinking Inference in {FPGA} Soft Logic}, 
    author={Wang, Erwei and Davis, James J. and Cheung, Peter Y. K. and Constantinides, George A.},
    booktitle=FCCM, 
    year={2019},
    month=apr,
    pages={26-34},
    doi={10.1109/FCCM.2019.00014}
}

@inproceedings{umuroglu2017finn,
	title        = {{FINN}: A Framework for Fast, Scalable Binarized Neural Network Inference},
	author       = {Umuroglu, Yaman and Fraser, Nicholas J. and Gambardella, Giulio and Blott, Michaela and Leong, Philip and Jahre, Magnus and Vissers, Kees},
	booktitle    = ACM,
	year         = 2017,
    month        = feb,
	series       = {FPGA '17},
	pages        = {65–74},
	doi          = {10.1145/3020078.3021744},
}

@book{b11,
    title = {{Design for Embedded Image Processing on {FPGAs}}},
    author = {Bailey, Donald G},
    publisher = {Wiley-IEEE Press},
    year = {2011},
    isbn = {9780470828502}
}

@article{kfmzm13,
    title = {{{FIR} filter optimization for video processing on {FPGAs}}},
    author = {Kumm, Martin and Fanghänel, Diana and Möller, Konrad and Zipf, Peter and Meyer-Baese, Uwe},
    journal = EURASIP_JORNAL_SIG,
    year = {2013},

    month = dec,
    number = {1},
    pages = {111},
    volume = {2013},
    doi = {10.1186/1687-6180-2013-111}
}

@book{p99,
    title = {{{VLSI} Digital Signal Processing Systems: Design and Implementation}},
    author = {Parhi, Keshab K.},
    publisher = {John Wiley \& Sons},
    year = {1999}
}

@book{de_dinechin2024application,
	title        = {Application-Specific Arithmetic},
	author       = {de Dinechin, Florent and Kumm, Martin},
	year         = 2024,
	publisher    = {Springer}
}

@inproceedings{habermann2022hardware,
	title        = {Hardware-Aware Quantization for Multiplierless Neural Network Controllers},
	author       = {Habermann, Tobias and Kühle, Jonas and Kumm, Martin and Volkova, Anastasia},
	booktitle    = APCCAS,
	year         = 2022,

    month        = nov,
	pages        = {541--545},
	doi          = {10.1109/APCCAS55924.2022.10090271}
}

@article{duarte2018fast,
	title        = {{Fast inference of deep neural networks in {FPGAs} for particle physics}},
	author       = {Duarte, J. and Han, S. and Harris, P. and Jindariani, S. and Kreinar, E. and Kreis, B. and Ngadiuba, J. and Pierini, M. and Rivera, R. and Tran, N. and Wu, Z.},
	journal      = JINST,
	year         = 2018,

    month        = jul,
	number       = {07},
	pages        = {P07027},
	volume       = 13,
	doi          = {10.1088/1748-0221/13/07/P07027}
}

@INPROCEEDINGS{andronic2023polylut,
    title={{PolyLUT}: Learning Piecewise Polynomials for Ultra-Low Latency {FPGA} {LUT}-based Inference}, 
    author={Andronic, Marta and Constantinides, George A.},
    booktitle=ICFPT, 
    year={2023},
    
    month=dec,
    pages={60-68},
    doi={10.1109/ICFPT59805.2023.00012}
}

@INPROCEEDINGS{resnet,
    title={Deep Residual Learning for Image Recognition}, 
    author={He, Kaiming and Zhang, Xiangyu and Ren, Shaoqing and Sun, Jian},
    booktitle=CVPR, 
    year={2016},

    month=jun,
    pages={770-778},
    doi={10.1109/CVPR.2016.90}
}

@INPROCEEDINGS{venieris2016fpgaconvnet,
    title={{fpgaConvNet}: A Framework for Mapping Convolutional Neural Networks on {FPGAs}}, 
    author={Venieris, Stylianos I. and Bouganis, Christos-Savvas},
    booktitle=FCCM, 
    year={2016},

    month=may,
    pages={40-47},
    doi={10.1109/FCCM.2016.22}
}

@misc{lou2024polylut,
	title        = {{PolyLUT-Add}: {FPGA-based} {LUT} Inference with Wide Inputs},
	author       = {Binglei Lou and Richard Rademacher and David Boland and Philip H. W. Leong},
	year         = 2024,
	note         = {{Under review}}
}

@misc{mobilenet,
      title={{MobileNets}: Efficient Convolutional Neural Networks for Mobile Vision Applications}, 
      author={Andrew G. Howard and Menglong Zhu and Bo Chen and Dmitry Kalenichenko and Weijun Wang and Tobias Weyand and Marco Andreetto and Hartwig Adam},
      year={2017},
      eprint={1704.04861},
      archivePrefix={arXiv},
}

@misc{dai2024kratos,
	title        = {{Kratos}: An {FPGA} Benchmark for Unrolled {DNNs} with Fine-Grained Sparsity and Mixed Precision},
	author       = {Xilai Dai and Yuzong Chen and Mohamed S. Abdelfattah},
	year         = 2024,
	eprint       = {2407.06033},
	archiveprefix = {arXiv},
	note         = {{Under review}}
}

@inproceedings{alexnet,
    title = {{ImageNet} Classification with Deep Convolutional Neural Networks},
    author = {Krizhevsky, Alex and Sutskever, Ilya and Hinton, Geoffrey E},
    booktitle = C_ACM,
    year = {2017},

    month = jun,
    pages = {84--90}
}

@ARTICLE{obj_det_survey,
    title={Object Detection in 20 Years: A Survey}, 
    author={Zou, Zhengxia and Chen, Keyan and Shi, Zhenwei and Guo, Yuhong and Ye, Jieping},
    journal={Proceedings of the IEEE}, 
    year={2023},
    month=mar,
    volume={111},
    number={3},
    pages={257-276},
    doi={10.1109/JPROC.2023.3238524}
}

@Article{yolov1to8,
    AUTHOR = {Hussain, Muhammad},
    TITLE = {{YOLO-v1} to {YOLO-v8}, the Rise of {YOLO} and Its Complementary Nature toward Digital Manufacturing and Industrial Defect Detection},
    JOURNAL = {Machines},
    YEAR = {2023},

    MONTH= jul,
    VOLUME = {11},
    NUMBER = {7},
    ARTICLE-NUMBER = {677},
    ISSN = {2075-1702},
    DOI = {10.3390/machines11070677}
}

@inproceedings{cnn_for_nlp,
    title = {Very Deep Convolutional Networks for Text Classification},
    author = {Conneau, Alexis and Schwenk, Holger and Barrault, Lo{\"\i}c and Lecun, Yann},
    booktitle = EACL,
    month = apr,
    year = 2017,
    pages = {1107--1116}
}

@ARTICLE{selfdriving_challanges,
    title={Edge Computing for Autonomous Driving: Opportunities and Challenges}, 
    author={Liu, Shaoshan and Liu, Liangkai and Tang, Jie and Yu, Bo and Wang, Yifan and Shi, Weisong},
    journal={Proceedings of the IEEE}, 
    year={2019},
    month=aug,
    volume={107},
    number={8},
    pages={1697--1716},
    doi={10.1109/JPROC.2019.2915983}
}

@ARTICLE{speechreq,
    title={Automatic Speech Recognition: Systematic Literature Review}, 
    author={Alharbi, Sadeen and Alrazgan, Muna and Alrashed, Alanoud and Alnomasi, Turkiayh and Almojel, Raghad and Alharbi, Rimah and Alharbi, Saja and Alturki, Sahar and Alshehri, Fatimah and Almojil, Maha},
    journal={IEEE Access}, 
    year={2021},
    volume={9},
    number={},
    month=sep,
    pages={131858-131876},
    doi={10.1109/ACCESS.2021.3112535}
}

@INPROCEEDINGS{augreality,
  title={State of the art of virtual reality technology}, 
  author={Anthes, Christoph and García-Hernández, Rubén Jesús and Wiedemann, Markus and Kranzlmüller, Dieter},
  booktitle=AERO, 
  year={2016},
  volume={},
  number={},
  pages={1-19},
  doi={10.1109/AERO.2016.7500674}}

@ARTICLE{tpu_eval,
    title={Motivation for and Evaluation of the First Tensor Processing Unit}, 
    author={Jouppi, Norman and Young, Cliff and Patil, Nishant and Patterson, David},
    journal={IEEE Micro}, 
    year={2018},
    month=may,
    volume={38},
    number={3},
    pages={10--19},
    doi={10.1109/MM.2018.032271057}
}

@INPROCEEDINGS{gpu_cnn_perf,
    title={Performance Analysis of GPU-Based Convolutional Neural Networks}, 
    author={Li, Xiaqing and Zhang, Guangyan and Huang, H. Howie and Wang, Zhufan and Zheng, Weimin},
    booktitle=ICPP, 
    year={2016},
    month=aug,
    volume={},
    number={},
    pages={67-76},
    doi={10.1109/ICPP.2016.15}
}

@MastersThesis{Zhenyu19MT,
	title={A Digits-Recognition Convolutional Neural Network on {FPGA}},
	author={Wang, Zhenyu},
	School= {Dept. of Electrical Engineering, Link\"oping University},
	year={2019},
	month = oct
}

@INPROCEEDINGS{cnns_review,
    title={A review on deep convolutional neural networks}, 
    author={Aloysius, Neena and Geetha, M.},
    booktitle=ICCSP, 
    year={2017},

    month=apr,
    volume={},
    number={},
    pages={0588--0592},
    doi={10.1109/ICCSP.2017.8286426}
}

@inproceedings{gao2023structural,
	title        = {Structural Alignment for Network Pruning through Partial Regularization},
	author       = {Gao, Shangqian and Zhang, Zeyu and Zhang, Yanfu and Huang, Feihu and Huang, Heng},
	year         = 2023,
    month        = oct,
	booktitle    = ICCV,
	pages        = {17356--17366},
	doi          = {10.1109/ICCV51070.2023.01596}
}

@inproceedings{gao2021network,
	title        = {Network Pruning via Performance Maximization},
	author       = {Gao, Shangqian and Huang, Feihu and Cai, Weidong and Huang, Heng},
	year         = 2021,
	booktitle    = CVPR,
	month        = jun,
	pages        = {9266--9276},
	doi          = {10.1109/CVPR46437.2021.00915}
}

@inproceedings{hossain2023computational,
	title        = {Computational Complexity Reduction Techniques for Deep Neural Networks: A Survey},
	author       = {Hossain, Md. Bipul and Gong, Na and Shaban, Mohamed},
	year         = 2023,
	booktitle    = AIBT,
    month        = sep,
	volume       = {},
	number       = {},
	pages        = {1--6},
	doi          = {10.1109/AIBThings58340.2023.10292477}
}

@inproceedings{wang2019haq,
	title        = {HAQ: Hardware-Aware Automated Quantization With Mixed Precision},
	author       = {Wang, Kuan and Liu, Zhijian and Lin, Yujun and Lin, Ji and Han, Song},
	year         = 2019,
	booktitle    = CVPR,
    month        = jun,
	pages        = {8604--8612},
	doi          = {10.1109/CVPR.2019.00881}
}

@inproceedings{miriyala2024mixed,
	title        = {Mixed Precision Neural Quantization with Multi-Objective Bayesian Optimization for on-Device Deployment},
	author       = {Miriyala, Srinivas S and Suhas, P K and Tiwari, Utsav and Rajendiran, Vikram N},
	year         = 2024,
	booktitle    = ICASSP,
    month        = apr,
	pages        = {6260--6264},
	doi          = {10.1109/ICASSP48485.2024.10448097}
}

@inproceedings{dong2021hao,
	title        = {HAO: Hardware-aware Neural Architecture Optimization for Efficient Inference},
	author       = {Dong, Zhen and Gao, Yizhao and Huang, Qijing and Wawrzynek, John and So, Hayden K.H. and Keutzer, Kurt},
	year         = 2021,
	booktitle    = FCCM,
    month        = may,
	pages        = {50--59},
	doi          = {10.1109/FCCM51124.2021.00014}
}

@inproceedings{andronic2024neuralut,
	title        = {{NeuraLUT}: Hiding Neural Network Density in Boolean Synthesizable Functions},
	author       = {Marta Andronic and George A. Constantinides},
	year         = 2024,
	month        = sep,
	booktitle    = FPL
}

@ARTICLE{main_ref,
  author={Li, Zhan and Zhang, Zhihan and Hu, Jie and Meng, Qunkang and Shi, Xingyu and Luo, Jun and Wang, Hao and Huang, Qijun and Chang, Sheng},
  journal={IEEE Transactions on Neural Networks and Learning Systems}, 
  title={A High-Performance Pixel-Level Fully Pipelined Hardware Accelerator for Neural Networks}, 
  year={2024},
  volume={},
  number={},
  pages={1-14},
  doi={10.1109/TNNLS.2024.3423664}
}

@inproceedings{flopoco,
  author = {de Dinechin, Florent},
  title = {Reflections on 10 years of {FloPoCo}},
  booktitle = {26th IEEE Symposium of Computer Arithmetic (ARITH-26)},
  year = {2019},
  month = jun,
  nopublisher = {IEEE}
}

@article{ImageNet,
Author = {Olga Russakovsky and Jia Deng and Hao Su and Jonathan Krause and Sanjeev Satheesh and Sean Ma and Zhiheng Huang and Andrej Karpathy and Aditya Khosla and Michael Bernstein and Alexander C. Berg and Li Fei-Fei},
Title = {{ImageNet Large Scale Visual Recognition Challenge}},
Year = {2015},
journal   = {International Journal of Computer Vision (IJCV)},
doi = {10.1007/s11263-015-0816-y},
volume={115},
number={3},
pages={211-252}
}

@inproceedings{lutmul_ref,
author = {Xie, Yanyue and Li, Zhengang and Diaconu, Dana and Handagala, Suranga and Leeser, Miriam and Lin, Xue},
title = {LUTMUL: Exceed Conventional FPGA Roofline Limit by LUT-based Efficient Multiplication for Neural Network Inference},
year = {2025},
isbn = {9798400706356},
publisher = {Association for Computing Machinery},
address = {New York, NY, USA},
doi = {10.1145/3658617.3697687},
booktitle = {Proceedings of the 30th Asia and South Pacific Design Automation Conference},
pages = {713–719},
numpages = {7},
location = {Tokyo, Japan},
series = {ASPDAC '25}
}

@article{finn_r,
author = {Blott, Michaela and Preu\ss{}er, Thomas B. and Fraser, Nicholas J. and Gambardella, Giulio and O’brien, Kenneth and Umuroglu, Yaman and Leeser, Miriam and Vissers, Kees},
title = {FINN-R: An End-to-End Deep-Learning Framework for Fast Exploration of Quantized Neural Networks},
year = {2018},
issue_date = {September 2018},
publisher = {Association for Computing Machinery},
address = {New York, NY, USA},
volume = {11},
number = {3},
issn = {1936-7406},
url = {https://doi.org/10.1145/3242897},
doi = {10.1145/3242897},
journal = {ACM Trans. Reconfigurable Technol. Syst.},
month = dec,
articleno = {16},
numpages = {23}
}

@ARTICLE{dcnn_acc,
  author={Huang, Wenjin and Luo, Conghui and Zhao, Baoze and Jiao, Han and Huang, Yihua},
  journal={IEEE Transactions on Neural Networks and Learning Systems}, 
  title={HCG: Streaming DCNN Accelerator With a Hybrid Computational Granularity Scheme on FPGA}, 
  year={2025},
  volume={36},
  number={10},
  pages={18681-18695},
  doi={10.1109/TNNLS.2025.3587694}}

@article{jsc_dataset,
doi = {10.1088/1748-0221/13/07/P07027},
url = {https://doi.org/10.1088/1748-0221/13/07/P07027},
year = {2018},
month = {jul},
publisher = {},
volume = {13},
number = {07},
pages = {P07027},
author = {Duarte, J. and Han, S. and Harris, P. and Jindariani, S. and Kreinar, E. and Kreis, B. and Ngadiuba, J. and Pierini, M. and Rivera, R. and Tran, N. and Wu, Z.},
title = {Fast inference of deep neural networks in FPGAs for particle physics},
journal = {Journal of Instrumentation},
}

@inproceedings{amigolut,
    author = {Weng, Olivia and Andronic, Marta and Zuberi, Danial and Chen, Jiaqing and Geniesse, Caleb and Constantinides, George A. and Tran, Nhan and Fraser, Nicholas J. and Duarte, Javier Mauricio and Kastner, Ryan},
    title = {Greater than the Sum of its LUTs: Scaling Up LUT-based Neural Networks with AmigoLUT},
    year = {2025},
    isbn = {9798400713965},
    publisher = {Association for Computing Machinery},
    address = {New York, NY, USA},
    url = {https://doi.org/10.1145/3706628.3708874},
    doi = {10.1145/3706628.3708874},
    booktitle = {Proceedings of the 2025 ACM/SIGDA International Symposium on Field Programmable Gate Arrays},
    pages = {25–35},
    numpages = {11},
    location = {Monterey, CA, USA},
    series = {FPGA '25}
}

@inproceedings{engels_work,
  author       = {S.E. Engel and P. Boutachkov and R. Singh},
  title        = {{Application of Machine Learning towards Particle Counting and Identification}},
  booktitle    = {Proc. IBIC'22},
  pages        = {508--511},
  eid          = {WEP42},
  language     = {english},
  venue        = {Kraków, Poland},
  series       = {International Beam Instrumentation Conference},
  number       = {11},
  publisher    = {JACoW Publishing, Geneva, Switzerland},
  month        = {12},
  year         = {2022},
  issn         = {2673-5350},
  isbn         = {978-3-95450-241-7},
  doi          = {10.18429/JACoW-IBIC2022-WEP42},
  url          = {https://jacow.org/ibic2022/papers/wep42.pdf},
}

@INPROCEEDINGS{10631128,
  author={Böttcher, Andreas and Kumm, Martin},
  booktitle={2024 IEEE 35th International Conference on Application-specific Systems, Architectures and Processors (ASAP)}, 
  title={Multiplier Design Addressing Area-Delay Trade-offs by using DSP and Logic resources on FPGAs}, 
  year={2024},
  volume={},
  number={},
  pages={217-225},
  doi={10.1109/ASAP61560.2024.00051}}

@INPROCEEDINGS{10579405,
  author={Böttcher, Andreas and Kumm, Martin},
  booktitle={2024 IEEE 31st Symposium on Computer Arithmetic (ARITH)}, 
  title={Small Logic-based Multipliers with Incomplete Sub-Multipliers for FPGAs}, 
  year={2024},
  volume={},
  number={},
  pages={124-131},
  doi={10.1109/ARITH61463.2024.00029}}

@inproceedings{DWN,
author = {Bacellar, Alan T. L. and Susskind, Zachary and Breternitz Jr., Mauricio and John, Eugene and John, Lizy K. and Lima, Priscila M. V. and Fran\c{c}a, Felipe M. G.},
title = {Differentiable weightless neural networks},
year = {2024},
publisher = {JMLR.org},
booktitle = {Proceedings of the 41st International Conference on Machine Learning},
articleno = {90},
numpages = {19},
location = {Vienna, Austria},
series = {ICML'24}
}

@inproceedings{TreeLUT,
author = {Khataei, Alireza and Bazargan, Kia},
title = {TreeLUT: An Efficient Alternative to Deep Neural Networks for Inference Acceleration Using Gradient Boosted Decision Trees},
year = {2025},
isbn = {9798400713965},
publisher = {Association for Computing Machinery},
address = {New York, NY, USA},
url = {https://doi.org/10.1145/3706628.3708877},
doi = {10.1145/3706628.3708877},
booktitle = {Proceedings of the 2025 ACM/SIGDA International Symposium on Field Programmable Gate Arrays},
pages = {14–24},
numpages = {11},
location = {Monterey, CA, USA},
series = {FPGA '25}
}

@inproceedings{andronic2025neuralut-assemble,
	author	= "Andronic, Marta and Constantinides, George A.",
	title		= "{NeuraLUT-Assemble: Hardware-Aware Assembling of Sub-Neural Networks for Efficient LUT Inference}",
	booktitle	= "{2025 IEEE 33rd Annual International Symposium on Field-Programmable Custom Computing Machines (FCCM)}",
	pages		= "208-216",
	publisher	= "IEEE",
	year		= 2025,
	note		= "doi: 10.1109/FCCM62733.2025.00077"
}

@INPROCEEDINGS{neural_LUT,
  author={Andronic, Marta and Constantinides, George A.},
  booktitle={2024 34th International Conference on Field-Programmable Logic and Applications (FPL)}, 
  title={NeuraLUT: Hiding Neural Network Density in Boolean Synthesizable Functions}, 
  year={2024},
  volume={},
  number={},
  pages={140-148},
  doi={10.1109/FPL64840.2024.00028}}

@misc{hls4ml,
      title={hls4ml: An Open-Source Codesign Workflow to Empower Scientific Low-Power Machine Learning Devices}, 
      author={Farah Fahim and Benjamin Hawks and Christian Herwig and James Hirschauer and Sergo Jindariani and Nhan Tran and Luca P. Carloni and Giuseppe Di Guglielmo and Philip Harris and Jeffrey Krupa and Dylan Rankin and Manuel Blanco Valentin and Josiah Hester and Yingyi Luo and John Mamish and Seda Orgrenci-Memik and Thea Aarrestad and Hamza Javed and Vladimir Loncar and Maurizio Pierini and Adrian Alan Pol and Sioni Summers and Javier Duarte and Scott Hauck and Shih-Chieh Hsu and Jennifer Ngadiuba and Mia Liu and Duc Hoang and Edward Kreinar and Zhenbin Wu},
      year={2021},
      eprint={2103.05579},
      archivePrefix={arXiv},
      primaryClass={cs.LG},
      url={https://arxiv.org/abs/2103.05579}, 
}

@article{scye17, 
year = {2017}, 
title = {{Efficient Processing of Deep Neural Networks: A Tutorial and Survey}}, 
author = {Sze, Vivienne and Chen, Yu-Hsin and Yang, Tien-Ju and Emer, Joel S}, 
journal = {Proceedings of the IEEE}, 
issn = {0018-9219}, 
doi = {10.1109/jproc.2017.2761740}, 
url = {http://ieeexplore.ieee.org/document/8114708/}, 
pages = {2295 -- 2329}, 
number = {12}, 
volume = {105}, 
month = {00}
}

\begin{IEEEbiography}[{\includegraphics[width=1in,height=1.25in, clip, keepaspectratio]{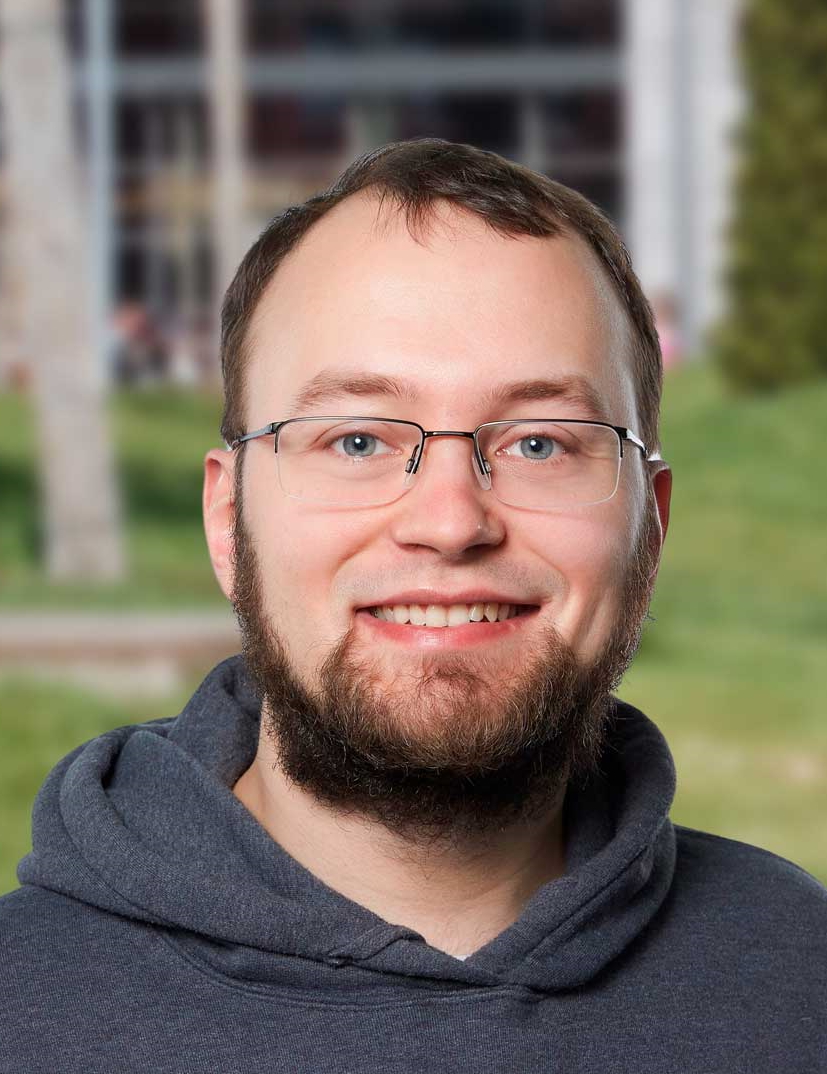}}]{Tobias Habermann} received the M.Sc. degree in applied computer science from the University of Applied Sciences Fulda, Germany, in 2020.

He is currently pursuing the Ph.D. degree in applied computer science under the supervision of Prof. Dr. Martin Kumm at the University of Applied Sciences, Fulda, Germany.

His current research interests include hardware-aware training of deep neural networks for FPGAs and the design of resource-efficient arithmetic for deep neural networks on FPGAs.
\end{IEEEbiography}

\begin{IEEEbiography}[{\includegraphics[width=1in,height=1.25in, clip, keepaspectratio]{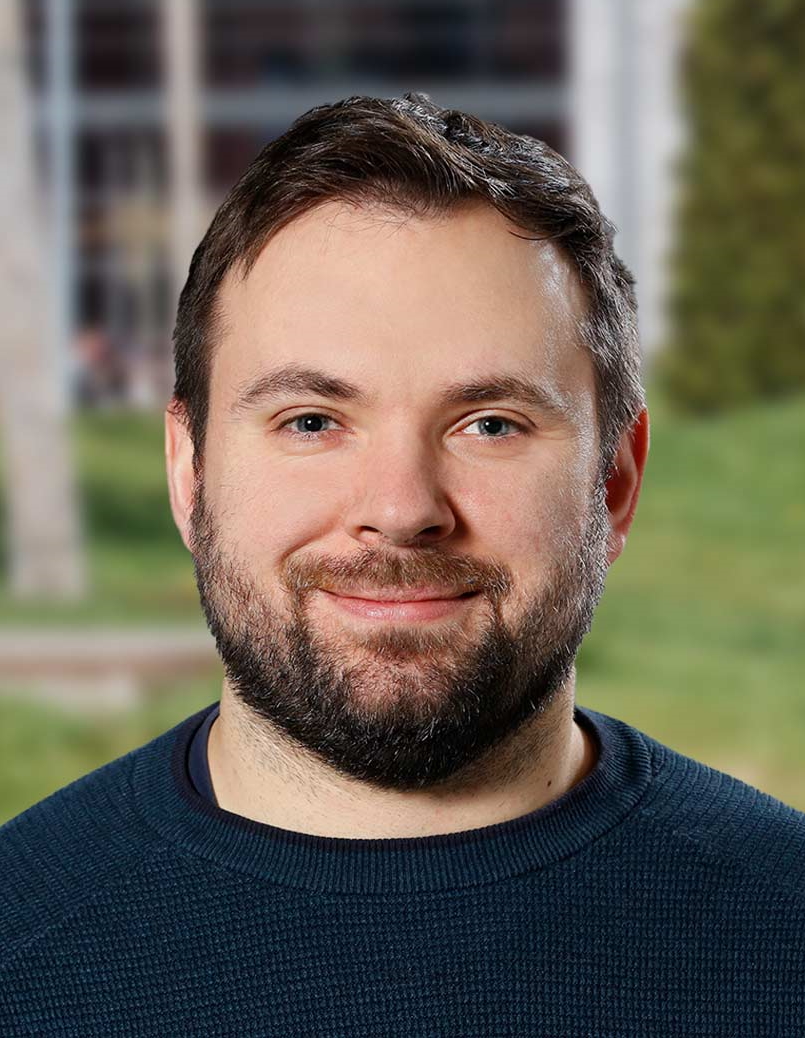}}]{Michael Mecik} received his M.Sc. degree in Applied Computer Science from the University of Applied Sciences Fulda, Germany, in 2017. After graduation, he worked in the field of embedded systems, focusing on IoT and healthcare devices. Currently, he is a laboratory engineer in embedded systems and robotics at Fulda University of Applied Sciences, Germany. He is pursuing a Ph.D. in Applied Computer Science under the supervision of Prof. Dr. Martin Kumm at Fulda University of Applied Sciences, Germany. His research interests include the optimization of deep neural networks for hardware accelerators.
\end{IEEEbiography}

\begin{IEEEbiography}[{\includegraphics[width=1in,height=1.25in, clip, keepaspectratio]{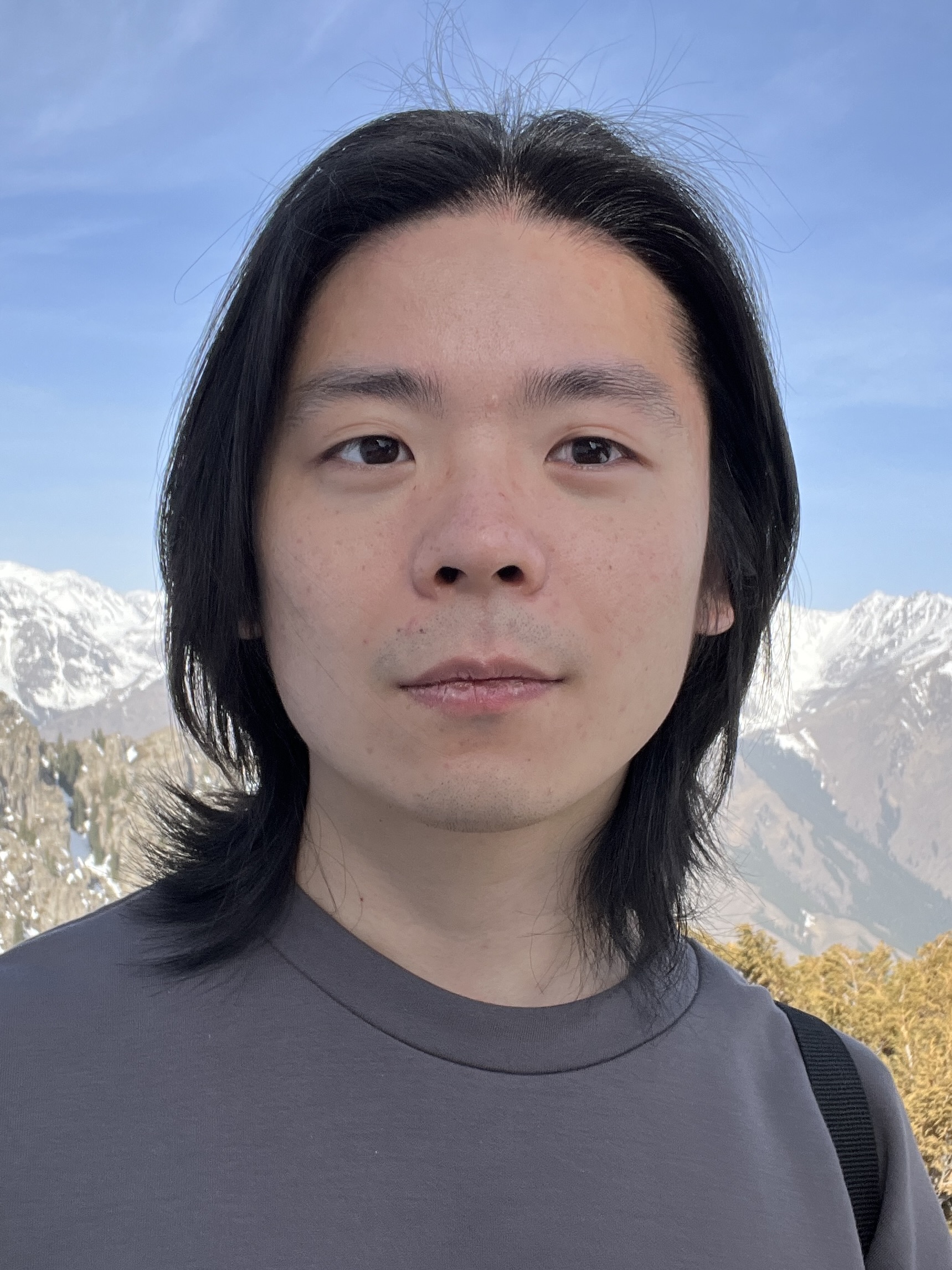}}]{Zhenyu Wang} received the M.Sc. degree in Electronics Engineering from Linköping University (LIU), Sweden, in 2019. He studied Electronics Engineering at LIU from 2017 to 2019 and moved back to China after graduation. From 2020 to present, he has worked as a digital IC design engineer in China Key System \& Integrated Circuit Co., Ltd. His research interests include the design of RF transceiver systems, the design of digital baseband, and low-power implementation of digital circuits.
\end{IEEEbiography}

\begin{IEEEbiography}[{\includegraphics[width=1in,height=1.25in, clip, keepaspectratio]{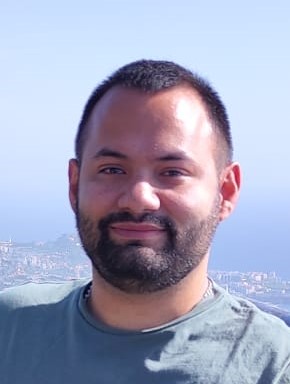}}]{César David Vera} received his B.Sc degree in Electronic Engineering from the Universidad Nacional Experimental del Táchira (UNET), Venezuela, in 2018. In 2019, he moved to Spain to study the Master's program in Electronic Systems Engineering at the Universidad Politécnica de Madrid (UPM) and got his M.Sc. degree in 2021. Since then, he has been working as an Embedded Firmware Engineer for Fossa Systems in Madrid. His interests include microcontrollers and FPGAs, as well as the design of algorithms for these devices. These algorithms are related to communication and codification, power-level-dependent activities and systems, synchronization between different devices, work threads, and hardware-sharing interactions.

\end{IEEEbiography}

\begin{IEEEbiography}[{\includegraphics[width=1in,height=1.25in,clip,keepaspectratio]{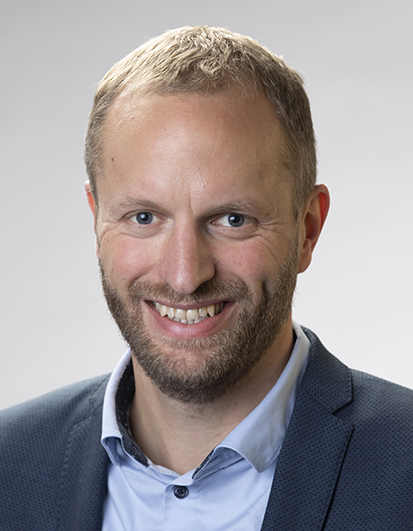}}]{Martin Kumm}
received the Dipl.-Ing. degree in electrical engineering from the Technical University of Darmstadt, Germany in 2007.

He was with GSI Darmstadt, working on digital control systems for particle accelerators from 2003 to 2009.
In 2015, he received his Ph.D. (\makebox{Dr.-Ing.}) degree from the University of Kassel, Germany.
He is currently a Professor for Embedded Systems at the Fulda University of Applied Sciences, Germany.
He has published over 60 research papers in international peer-reviewed journals, conferences, and workshops.
His work received best paper awards at the
IEEE International Symposium on Computer Arithmetic (ARITH), the
IEEE International Symposium on Field-Programmable Custom Computing Machines (FCCM), and the IEEE International Symposium on Defect and Fault Tolerance in VLSI and Nanotechnology Systems (DFT).
His research interests are application-specific arithmetic and its optimization, as well as architectures for deep learning with particular emphasis on reconfigurable systems.
\end{IEEEbiography}

\begin{IEEEbiography}
[{\includegraphics[width=1in,height=1.25in, clip,keepaspectratio]{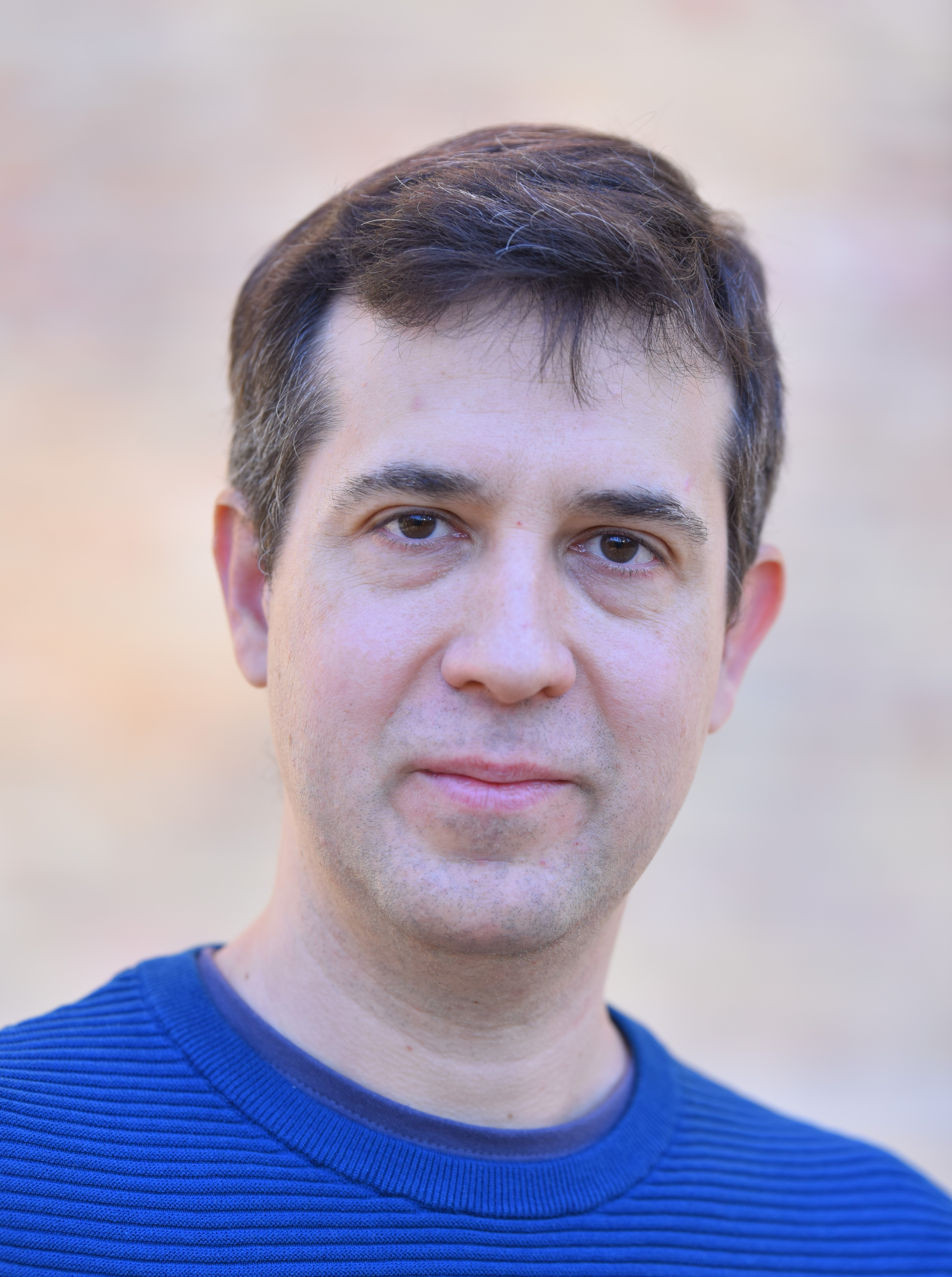}}]{Mario Garrido} (Senior Member, IEEE) received the M.Sc. and Ph.D. degrees in Telecommunications Engineering from Universidad Politécnica de Madrid (UPM), Spain, in 2004 and 2009, respectively.

In 2010 he moved to Sweden to work as a Post-Doctoral Researcher with the Department of Electrical Engineering, Link\"oping University. From 2012 to 2019 he was an Associate Professor with the Department of Electrical Engineering, Link\"oping University. In 2019 he returned to UPM with a Ram\'on y Cajal Research Fellowship and since 2024 he has been an Associate Professor with the Department of Electronic Engineering at UPM. So far, he has been the author of more than 50 scientific publications. His research interests include optimized hardware design for signal-processing applications, design of hardware architectures for the fast Fourier transform (FFT), circuits for data management, the CORDIC algorithm, neural networks, and circuits to calculate statistical and mathematical operations. His research covers high-performance circuits for real-time computation and designs for small area and low-power consumption. He appeared in the "World's Top 2\% Scientists List" elaborated by Standford University in 2022, 2023, 2024, and 2025.  \end{IEEEbiography}

\end{document}